\documentclass[manuscript,screen, pbalance]{acmart}

\AtBeginDocument{%
  }

\setcopyright{acmlicensed}
\copyrightyear{2026}
\acmYear{2026}
\acmDOI{XXXXXXX.XXXXXXX}
\acmJournal{TOSEM}
\acmVolume{1}
\acmNumber{1}
\acmArticle{1}

\usepackage{graphicx}
\usepackage{listings}
\usepackage{etc}
\usepackage{tcolorbox}
\usepackage{enumitem}
\usepackage{adjustbox}
\usepackage{minted}
\usepackage{algorithm}
\usepackage{algorithmic}

\setcopyright{none}
\renewcommand\footnotetextcopyrightpermission[1]{} 
\definecolor{bg}{rgb}{0.95,0.95,0.95}

\definecolor{codegreen}{HTML}{3F6B4F}
\newtcolorbox{rqbox}[1][]{
    colback=green!10, colframe=codegreen, arc=1mm, boxrule=0.5pt,
    coltitle=white, fonttitle=\bfseries, title=#1,
    top=0.5mm, bottom=0.5mm, left=1mm, right=1mm,
    toptitle=0.5mm, bottomtitle=0.5mm
}

\newtcolorbox{findingbox}[1][]{
    colback=green!10, colframe=codegreen, arc=1mm, boxrule=0.5pt,
    coltitle=white, fonttitle=\bfseries, title=#1,
    top=0.5mm, bottom=0.5mm, left=1mm, right=1mm,
    toptitle=0.5mm, bottomtitle=0.5mm
}

\usepackage{listings,xcolor}
\usepackage{mdframed}

\definecolor{promptbg}{HTML}{EDF3E7}     
\definecolor{promptframe}{HTML}{BFD4B0}  
\definecolor{promptslot}{HTML}{2F6B3D}   
\definecolor{prompttext}{HTML}{2B332B}   

\lstdefinestyle{prompt}{
  basicstyle=\ttfamily\small\color{prompttext},
  backgroundcolor=\color{promptbg},
  frame=single, rulecolor=\color{promptframe}, framerule=0.8pt,
  breaklines=true, columns=fullflexible,
  xleftmargin=0.8em, xrightmargin=0.8em,
  framexleftmargin=0.5em, framexrightmargin=0.5em,
  aboveskip=0.9em, belowskip=0.9em,
  moredelim=[s][\color{promptslot}\bfseries]{\{}{\}},
}

\mdfdefinestyle{sagealg}{
  backgroundcolor=promptbg, linecolor=promptframe, linewidth=0.8pt,
  roundcorner=2pt, innertopmargin=4pt, innerbottommargin=4pt,
  innerleftmargin=6pt, innerrightmargin=6pt,
}

\definecolor{cbBg}{HTML}{EDF3E7}      
\definecolor{cbFrame}{HTML}{BFD4B0}   
\definecolor{cbText}{HTML}{1A1A1A}    
\definecolor{cbNum}{HTML}{9CA3AF}     
\definecolor{cbComment}{HTML}{6B7280} 
\definecolor{cbKw}{HTML}{6D28D9}      
\definecolor{cbType}{HTML}{0E7490}    
\definecolor{cbStr}{HTML}{4D7C0F}     
\definecolor{cbDir}{HTML}{7C3AED}     

\lstdefinestyle{cppbox}{
  language=C++,
  backgroundcolor=\color{cbBg},
  basicstyle=\ttfamily\scriptsize\color{cbText},
  deletekeywords=[1]{int,long,short,char,bool,float,double,void,auto,unsigned,signed,size_t},
  keywordstyle=\color{cbKw}\bfseries,
  keywords=[2]{int,long,short,char,bool,float,double,void,auto,unsigned,signed,size_t},
  keywordstyle=[2]\color{cbType},
  stringstyle=\color{cbStr},
  commentstyle=\color{cbComment}\itshape,
  directivestyle=\color{cbDir},
  numbers=left, numberstyle=\tiny\color{cbNum}, numbersep=9pt,
  frame=single, rulecolor=\color{cbFrame}, framerule=0.6pt,
  xleftmargin=16pt, xrightmargin=4pt, framexleftmargin=2pt,
  aboveskip=0.9em, belowskip=0.7em,
  showstringspaces=false, breaklines=true, tabsize=2,
  columns=fullflexible, keepspaces=true, upquote=true,
}

\begin{document}

\title{\textbf{Beyond the \textit{Need for Speed}}: Energy-Aware Code Generation via Simulation-Guided Reinforcement Learning}


\author{Saurabhsingh Rajput, Tushar Sharma}
\affiliation{
  \institution{Dalhousie University}
  \city{Halifax}
  \country{Canada}}
\email{{saurabh, tushar}@dal.ca}

\renewcommand{\shortauthors}{Rajput and Sharma}

\begin{abstract}
Code models strictly prioritize functional correctness, leaving software energy efficiency as an unoptimized byproduct. Training models to generate energy-efficient code requires reproducible feedback at scale, which physical hardware measurement cannot reliably provide due to variance.

In this paper, we replace hardware profiling with a deterministic architectural simulation harness to build \textbf{Green Tea}, a corpus of $3.5$ million evaluations across $1{,}474$ C++ problems. We train an energy-aware code model via supervised fine-tuning on energy-contrastive pairs, followed by closed-loop reinforcement learning (GRPO) using simulation-in-the-loop feedback. To rigorously evaluate deployment readiness, we introduce the \emph{\textbf{C}orrectness-\textbf{A}djusted \textbf{R}eduction in \textbf{E}nergy \textbf{T}otal} (CARET), a metric that explicitly penalizes code that sacrifices functionality for efficiency. 

On $143$ held-out problems, our simulation-in-the-loop pipeline achieves $12.63\%$ CARET, nearly tripling the gain of fine-tuning alone, and successfully beats the energy efficiency of human-expert references on $58.4\%$ of its valid outputs. Furthermore, our analysis exposes the \emph{IPC trap}: standard throughput proxies like Instructions-Per-Cycle (IPC) actively misrank true energy efficiency on $67.8\%$ of problems, proving the absolute necessity of direct energy simulation. By releasing our dataset and infrastructure, we bypass the $263{,}000$ CPU-hours required for reproduction, structurally empowering the community to deploy inherently energy-efficient code generation models.
\end{abstract}



\keywords{energy-efficient code, large language models, reinforcement learning, simulation-guided training, software energy consumption}

\maketitle

\section{Introduction}
\label{sec:introduction}
AI systems now produce a large and rising portion of the world's software. That shift raises energy consumption on two distinct fronts: the massive compute required to train and run the AI models themselves, and the execution energy of the generated code every time it runs on downstream hardware. We address the second front, where the operational energy footprint is far from fixed because software implementation choices dictate the underlying hardware workload. Implementations of the same task differ by up to two orders of magnitude in energy, depending on language, algorithm, \textsc{API} and data-structure choices~\cite{pereira2021ranking,rajput2024enhancing,rajput2026energy}. Hardware efficiency keeps improving, but historical evidence suggests that hardware gains alone expand aggregate consumption rather than reduce it, a tendency consistent with the Jevons paradox~\cite{IEA2025EnergyAI}. Software efficiency is the complementary lever, because a program's energy is set by the work its code makes the hardware do.
Yet the authorship of that software is shifting. Code production is moving from human-written to AI-assisted to AI-generated~\cite{cui2025effects,shen2026ai}, and industry analysts forecast that most enterprise software engineering work will involve AI assistants by the end of the decade~\cite{Gartner2024AI}. These models are trained for functional correctness rather than resource efficiency. Because a program's energy consumption depends on how its compiled code exercises the processor, a code model that never sees that hardware signal tends to emit \emph{hardware-oblivious} code that is correct but inefficient~\cite{cursaru2024controlled,vartziotis2024learn}. This algorithmic inefficiency compounds systematically over successive model generations. As hardware-oblivious code accumulates in the public corpora that train the next generation of models, these same inefficient patterns are learned and amplified in a `\textit{garbage-in, garbage-out}' feedback loop, much as recursive training on generated data triggers \emph{model collapse} and degrades other capabilities~\cite{shumailov2024ai}. With data-center electricity demand projected to rise by roughly $15\%$ per year through the decade~\cite{IEA2025EnergyAI}, the training \emph{signal} that shapes how models generate code, meaning the objective they are optimized to improve, is a direct lever on the energy cost of software.

Three existing lines of work come closest to supplying that signal, yet each falls short of it. Manual energy profiling, surveyed in green software engineering reviews~\cite{verdecchia2022green}, demands deep domain expertise and does not scale to the volume of AI-generated code. Green prompting prepends energy-efficiency instructions to an instruction-tuned model~\cite{cursaru2024controlled,vartziotis2024learn}, but it assumes that a model trained for correctness can recognize energy-efficient patterns at inference, which current code models do not do reliably. Execution-feedback reinforcement learning (RL) rewards program-level outcomes, yet existing systems reward correctness alone. CodeRL~\cite{le2022coderl} and AlphaCode~\cite{li2022alphacode} ignore energy entirely, while the Performance-Improving Edits (PIE) corpus~\cite{shypula2023pie} targets runtime instead. In short, \textit{no existing method provides a direct energy signal at the throughput required for model training}, leaving the field reliant on the untested assumption that faster code is inherently more energy-efficient.

That assumption is the first of two obstacles. Across programming tasks of different complexity classes, runtime and energy move together, so a faster solution is usually a lower-energy one. That agreement rests on average power being similar across the solutions being compared, which holds across different tasks but not within a single problem. Energy is the product of execution time and power draw, so two solutions that finish in the same number of cycles can still differ in energy when their instruction mix changes the power drawn per cycle. A model trained to minimize runtime then optimizes the wrong quantity on the solutions that run for similar times, and a training pair labeled by runtime can rank the two solutions the wrong way for energy. This within-problem gap has not been measured at scale. The second obstacle is measurement. The standard CPU energy interface, Intel's Running Average Power Limit (RAPL)~\cite{garcia2019exploration}, updates its on-chip counters only every few milliseconds, so on sub-second programs the readings are noisy and rank such programs unreliably~\cite{hahnel2012rapl}. RAPL also measures one power domain at a time and cannot produce the tens of thousands of energy values an RL run consumes, and external power meters are no better, since they share that throughput limit and report whole-system energy rather than the energy of the program alone. \textit{No existing method meets all three needs at once}, namely a signal that targets energy directly, evidence that the signal ranks solutions correctly, and a reproducible reward fast enough for large-scale training.

To close these gaps, we train a code model to optimize a program's energy directly for a target hardware architecture. Given an inefficient program, the model then produces a more efficient refactoring by default, without depending on an inference-time prompt or a runtime proxy. The enabling idea is deterministic architectural simulation~\cite{binkert2011gem5,carlson2014sniper}, a standard and rigorous technique widely used in computer architecture research and hardware design to model exactly how software exercises a processor. By computing energy from this software model, simulation produces repeatable per-program energy measurements at a throughput that physical hardware profiling cannot reach, which is what makes training against an energy signal feasible. The procedure has two stages. The first fine-tunes the model on energy-contrastive pairs from \textbf{Green Tea}, our corpus of C++ programs in which every program carries simulation-derived labels for its energy, cycle count, and instruction mix. Each pair couples a less efficient program with a more efficient one for the same task. This contrastive pairing allows the model to implicitly learn the source-level patterns that separate efficient implementations from inefficient ones at fixed task complexity, inherently discovering energy-saving transformations from the data~\cite{mehditabar2026validated}. The second stage applies reinforcement learning under a simulated reward computed from each refactoring's energy and runtime. The model proposes refactoring candidates, the simulator compiles, runs, and measures each one, and that measurement becomes the reward that updates the model. We call this the \emph{simulation-in-the-loop} cycle. Unlike fine-tuning on fixed pairs, it gives the model feedback on whether its own output compiles, passes tests, and actually saves energy when run.

To measure outcomes without rewarding efficient but incorrect code, we introduce the \emph{\textbf{C}orrectness-\textbf{A}djusted \textbf{R}eduction in \textbf{E}nergy \textbf{T}otal} (CARET), which credits an output's energy saving in proportion to the fraction of tests it passes and assigns zero to code that does not compile. On a held-out benchmark of $143$ PIE problems~\cite{shypula2023pie} unseen during training, the full procedure reaches a CARET of $12.63\%$, a $2.84\times$ gain over fine-tuning alone, while compiling $81.7\%$ of its outputs and producing a more efficient program than the human-written reference on $58.4\%$ of its valid outputs. An evaluation over the fine-tuning initialization and the reward signal keeps all five configurations within a $4.22\,$pp CARET band, each above the strongest fine-tuning-only variant. Two findings stand out. The simulation-in-the-loop cycle, not any single reward or initialization, drives the aggregate energy reduction. The reward choice drives how often the model beats the human reference, where the two energy-delay reward configurations roughly double that rate over the energy-only and runtime rewards. To summarize, we make four contributions with this study:

\begin{enumerate}
\item \textbf{Simulation-in-the-loop reinforcement learning for energy-aware code optimization.} A training procedure that rewards a code model with a reproducible simulated reward computed from the energy and runtime of its own generated refactorings, evaluated for every rollout.

\item \textbf{A within-problem characterization of the energy-optimization landscape.} An analysis of $2.6$ million solution pairs across $1{,}474$ programming problems that quantifies how often a throughput proxy misranks energy within a problem, most severely instructions per cycle, an effect we name the \emph{IPC trap} (occurring in $67.8\%$ of problems). We also report the source-level transformation classes the trained model uses and the conditions under which simulation supplies ranking information that runtime cannot.

\item \textbf{The Green Tea dataset and a reproducible benchmark.} A corpus of $3{,}507{,}435$ deterministic energy simulations over $1{,}474$ C++ problems, with $12{,}455$ energy-contrastive training pairs and a $143$-problem held-out benchmark. The dataset represents $263{,}000$ CPU-hours of simulation, so releasing its pre-computed labels lets others reproduce the procedure without re-incurring that cost.

\item \textbf{A scalable energy-simulation training harness.} We provide the full containerized infrastructure linking the LLM, the compiler, and the architectural simulator (Sniper/McPAT). This enables researchers to evaluate and train models for energy efficiency at scale without building complex hardware-simulation pipelines from scratch.
\end{enumerate}

\textbf{Replication.} The Green Tea dataset, the $143$-problem benchmark, the training and evaluation code, and the trained models are publicly available in our replication package.\footnote{The complete replication package (including datasets, models, and code) is available at: \url{https://github.com/SMART-Dal/green-tea}}

The paper is organized as follows: Section~\ref{sec:related} covers related work; Section~\ref{sec:overview} defines the study design; Section~\ref{sec:methods} details the methodology; Section~\ref{sec:results} answers the research questions; Section~\ref{sec:discussion} synthesizes findings; Section~\ref{sec:implication} extracts implications; and Sections~\ref{sec:threats}--\ref{sec:conclusion} address threats and conclusions.
\section{Background and Related Work}
\label{sec:related}
This section places our work in context. We first survey the magnitude of software-level energy variation that makes optimization worthwhile. We then describe the two ways to measure a program's energy, hardware profiling and architectural simulation, and explain why simulation can supply an energy signal at the scale training needs. We close with learned approaches to code optimization and reinforcement learning for code, which together frame the gap this work addresses.

\subsection{Software Energy Variation}
\label{sec:related-energy}

Energy consumption varies by orders of magnitude across implementations of the same task. Pereira et al.~\cite{pereira2021ranking} measure $27$ programming languages on $10$ benchmark problems and report normalized energy ratios approaching $76\times$ between the most and least energy-efficient languages. Within a single language, algorithm and data-structure choices produce variation of a similar scale~\cite{shypula2023pie}. This headroom motivates automated optimization, but the optimization signal must stay reliable at the scale at which automation operates. Compiler optimization flags automate intra-algorithmic improvements such as loop unrolling, vectorization, and register allocation~\cite{aho2007compilers}, yet they cannot discover the algorithmic transitions, for example $O(n^2) \to O(n \log n)$, that source-level rewriting reaches. Auto-tuners for flag selection and parameter search likewise operate within a fixed source program rather than rewriting it. Green software engineering surveys~\cite{verdecchia2022green,mehditabar2026validated} catalog energy-aware coding patterns but rely on manual expertise, and recent studies that prompt code models for more energy-efficient output~\cite{cursaru2024controlled,vartziotis2024learn} report small and inconsistent effects.

\subsection{Hardware Energy Profiling}
\label{sec:related-measurement}
One way to obtain a program's energy is direct hardware profiling, which reads energy from the running processor. Intel's Running Average Power Limit (RAPL)~\cite{garcia2019exploration} is the standard on-chip interface on x86 processors and reports the energy drawn by a whole processor package, while external power meters read the energy the entire system draws from the wall socket. Hardware profiling captures the workload as it actually runs, but the readings are subject to thermal drift, scheduler interference, and quantization, the rounding of energy into discrete counter steps. RAPL updates its counters only on the order of milliseconds, so a program that runs for less than one update interval is measured with large timing artifacts, and reported run-to-run variance reaches $5$ to $15\%$ on sub-second workloads~\cite{hahnel2012rapl}. Ranking two short programs by such readings is therefore unreliable. Hardware profiling also measures one power domain at a time and does not scale to the tens of thousands of evaluations that large-scale model training consumes. Shypula et al.~\cite{shypula2023pie} reduce hardware timing variance for runtime with a custom benchmarking harness, but runtime needs only a relative ordering, whereas an energy reward needs deterministic, repeatable values.

\subsection{Architectural Energy Simulation}
\label{sec:related-simulation}

The alternative is to estimate energy from a software model of the processor, an approach called \emph{architectural simulation}. Here \emph{architecture} refers to the hardware, the microarchitecture of the CPU, rather than the software architecture of a program. A simulator runs the program against this model and records what the processor does. Because the model is deterministic, it returns the same measurement on every run, at the cost of running far slower than native execution. Although a simulator necessarily targets one specific processor architecture, recent empirical evidence shows that the energy performance of code generated by language models remains highly correlated across different hardware platforms, suggesting that simulation-derived optimizations are inherently portable~\cite{solovyeva2025ai}. Three classes trade detail against speed differently. \emph{Cycle-accurate simulators} such as gem5~\cite{binkert2011gem5} model every pipeline stage and support diverse instruction sets, but run at roughly $200{,}000$ instructions per second, so building a large simulation corpus with them would take years of compute. \emph{Interval simulators} such as Sniper~\cite{carlson2014sniper} and ZSim~\cite{sanchez2013zsim} abstract the out-of-order core at region boundaries rather than cycle by cycle, which makes them orders of magnitude faster on application-level workloads while remaining accurate. On integer workloads Sniper reports a lower mean absolute percentage error in instructions per cycle (IPC) than gem5, $17.6\%$ against $37.1\%$ under the outlier-excluded protocol of that comparison~\cite{x86simcomparison2016}. \emph{Static analyzers} such as LLVM-MCA~\cite{llvmmca} infer throughput from instruction-scheduling tables without executing the program, which makes them unsuitable for data-dependent workloads where branch behavior and cache-miss rates depend on the input. GPU power simulators target accelerator workloads and do not apply to single-threaded CPU programs.

A simulator that reports cycles is enough for runtime, but energy also requires translating per-component activity into joules, and McPAT~\cite{dong2010mcpat} performs that step. It reads \emph{activity counters} from a host simulator, the counts of events such as cache accesses, branch mispredictions, arithmetic-logic-unit operations, and memory-controller activity, and maps them to energy, power, and area estimates through a parameterized processor model. Because the mapping is linear in the activity counts, the energy ordering between two implementations is preserved under a global calibration factor. Sniper, gem5, and McPAT are established infrastructure in computer-architecture research, and Sniper integrates natively with McPAT, so the pair produces deterministic per-program energy fast enough for the volume that large-scale training consumes. ZSim reaches a similar speed but has no native energy model, and gem5 is far slower. A learned cost model that predicts throughput could estimate energy faster still, but such a model needs per-target training data and does not preserve energy rankings by construction, whereas the activity-to-energy mapping is rank-preserving without any learned surrogate.

\subsection{Learned Approaches to Code Optimization}
\label{sec:related-llm}

Large language models generate code effectively~\cite{chen2021codex,li2022alphacode,roziere2023codellama,hui2024qwen25coder,guo2024deepseekcoder}, drawing heavily on massive open-source datasets and execution traces~\cite{grossman_compile_2024}, but standard training optimizes for functional correctness rather than resource efficiency. A line of work narrows that gap. PIE~\cite{shypula2023pie} pairs slow and fast competitive-programming solutions and fine-tunes a code model with score-conditioned prompts for runtime speedup. More recently, Afterburner~\cite{du2025afterburner} applies reinforcement learning to improve code efficiency. Supersonic~\cite{Chen2024} trains a compact sequence-to-sequence model that emits a source-level diff to speed up a C/C++ program. ECCO~\cite{waghjale2024} is a reproducible benchmark that evaluates the runtime and memory efficiency of generated Python programs under execution-based measurement. These approaches optimize for time or memory, but none directly target energy. This omission is critical, as previous work establishes that runtime is not a universally reliable proxy for energy consumption; depending on the hardware context and instruction mix, a reduction in runtime does not necessarily guarantee a reduction in energy~\cite{weber2023twins,rua2024large,pereira2021ranking,rajput2026flipflop}. Furthermore, recent empirical studies profiling the energy footprint of code generated by large language models~\cite{solovyeva2025ai,ashraf2025energyaware} reinforce the urgent need for active, energy-aware training pipelines.

EffiBench~\cite{huang2024effibench} normalizes the execution time and peak memory of generated Python programs against a canonical solution, while Mercury~\cite{du2024mercury}, EvalPerf~\cite{liu2024evalperf}, and ENAMEL~\cite{qiu2025enamel} rigorously evaluate and score runtime percentile speedups over complex competitive-programming sets. All of these approaches gate efficiency on binary correctness, crediting only outputs that pass every test. CodeCarbon~\cite{codecarbon} and MLCO2~\cite{lacoste2019quantifying} estimate the energy of \emph{training} a model rather than the energy of the code a model generates at inference. Three gaps run across these efforts. None measures the energy of generated code at inference, none derives its measure from a deterministic source rather than wall-clock timing, and all discard an output that fails any single test rather than crediting its partial correctness. Closing these gaps calls for an efficiency measure built on energy, determinism, and graded correctness, which we develop in Section~\ref{sec:setup}.

Energy is not a relabeling of those dimensions. Two implementations of the same problem can run in the same number of cycles yet draw different power, so a model tuned for runtime or memory can leave energy unimproved or even raise it. Optimizing energy therefore requires a signal that ranks implementations by joules, and producing that signal at training scale is the obstacle hardware measurement cannot clear. Deterministic architectural simulation can supply it, which is the step this line of work has not taken.

\subsection{Reinforcement Learning for Code Generation}
\label{sec:related-rl}

\emph{Reinforcement learning} (RL) trains a model, the \emph{policy}, by having it act and then rewarding the actions that lead to good outcomes. In code generation the policy is the code model, an action is a program the model produces, and a \emph{rollout} is one such generated program together with the \emph{reward} it earns, the single scalar the model is trained to maximize. Training is \emph{on-policy} when the rewarded rollouts come from the current model rather than from a fixed dataset, so the model learns from its own latest outputs. What the reward measures therefore decides what the model learns to optimize.

Execution-based rewards use this mechanism to improve functional correctness. The reward comes from running the generated program against tests or a judge, which supplies a signal that token-level supervised training cannot. CodeRL~\cite{le2022coderl} uses unit-test outcomes as the reward, and AlphaCode~\cite{li2022alphacode} uses competitive-programming judge feedback. PPOCoder~\cite{shojaee2023ppocoder} adds compiler and structure-matching signals to the reward, RLTF~\cite{liu2023rltf} makes the reward fine-grained by locating the failing test, and StepCoder~\cite{dou2024stepcoder} eases exploration by training through a curriculum of code-completion subtasks. Recent methods like SWE-RL~\cite{wei2025swerl} and RLEF~\cite{gehring2025rlef} demonstrate that execution feedback scales to open software-evolution environments, while systems like DRIVE~\cite{zhu2025drive} standardize best practices for verifiable rewards. Across all these systems the reward strictly encodes correctness, and none considers energy or any other resource cost.
\section{Study Design}
\label{sec:overview}
\begin{figure}[t]
\centering
\includegraphics[width=\textwidth]{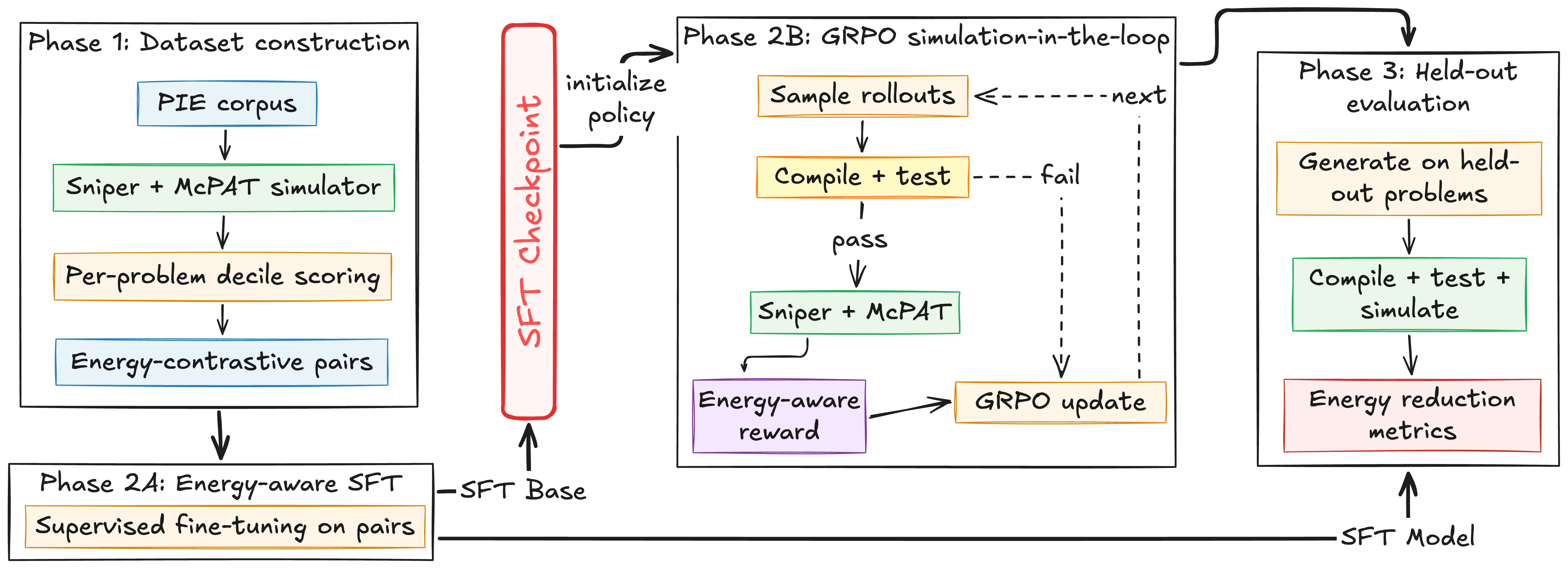}
\caption{Overview of the three-phase procedure. Phase~1 builds the Green Tea dataset by simulating C++ solutions and pairing a less efficient baseline with a more efficient target from the same problem. Phase~2 trains the model in two stages. Phase~2A fine-tunes the base model on these pairs, and Phase~2B closes the simulation-in-the-loop cycle, rewarding each generated refactoring by its simulated energy and runtime. Phase~3 evaluates the trained model on held-out problems.}
\label{fig:overview}
\end{figure}

The goal of this study is to determine whether a code model can be trained to optimize a program's energy consumption while preserving its behavior.
We pursue this goal in three phases, summarized in Figure~\ref{fig:overview}. The first phase builds the \textbf{Green Tea} dataset. We simulate a large corpus of C++ solutions, rank the solutions to each problem by their simulated energy, and pair a less efficient solution with a more efficient one for the same problem, which gives the energy-contrastive pairs the model trains on. The second phase trains the model in two stages that differ in where their signal comes from. Fine-tuning learns from this fixed set of pairs, and reinforcement learning then learns from the model's own generated refactoring candidates and how much energy each one saves. The final phase evaluates the trained model on held-out problems it never saw during training.

We design four research questions to evaluate the proposed approach. \textit{RQ1} examines the optimization signal that both stages rely on. \textit{RQ2} evaluates the first stage in isolation. \textit{RQ3} evaluates the second stage and asks whether the simulation-in-the-loop cycle, the reward signal, or the fine-tuning initialization is responsible for the gain. \textit{RQ4} characterizes the source-level transformations the trained model produces. Each question examines the relevance and efficacy of one part of the procedure, answered in detail in Section~\ref{sec:results}.

\begin{rqbox}
\textit{\textbf{RQ1:} Do standard throughput proxies reliably rank energy efficiency within a single problem, and where do they diverge?}
\end{rqbox}

Prior work on learned code optimization uses runtime as a proxy for energy, assuming the two track each other closely~\cite{shypula2023pie,rajput2026energy}. While this assumption holds across disparate complexity classes, it fractures within a single problem where cycle counts cluster tightly. In this regime, remaining energy differences stem from instruction-mix power variations, rendering runtime blind. Identifying where runtime and energy diverge determines the required training infrastructure, because a proxy that mis-ranks energy biases every training pair built from it.

\begin{rqbox}
\textit{\textbf{RQ2:} Can supervised fine-tuning alone produce a deployable energy-aware code model?}
\end{rqbox}
Supervised fine-tuning is the most direct way to teach a code model the source-level patterns that separate an efficient implementation from an inefficient one. However, because fine-tuning never executes the code it learns from, it receives zero feedback on whether an output compiles, passes its tests, or actually saves energy when it runs. This question asks whether learning from fixed examples suffices, or if a deployable optimizer strictly requires dynamic execution feedback from its own outputs.

\begin{rqbox}
\textit{\textbf{RQ3:} Does on-policy reinforcement learning under a deterministic simulation reward improve on static supervision, and if so, is the improvement robust to the choice of reward signal and fine-tuning initialization?}
\end{rqbox}

On-policy reinforcement learning allows the model to learn from its own outputs, correcting code that looks superficially right but fails to compile, pass its tests, or save energy. This question isolates whether this closed-loop feedback mathematically improves upon fine-tuning, and systematically disentangles whether those gains originate from the reward signal, the fine-tuning initialization, or the simulation-in-the-loop cycle itself.

\begin{rqbox}
\textit{\textbf{RQ4:} What kind of transformations does the trained model produce, and do they reach the cases where runtime proxies fail?}
\end{rqbox}

A model might achieve energy reductions through profound algorithmic shifts, structural constant-factor optimizations, or trivial cosmetic edits that change the source without changing what it computes or costs. Classifying these transformations is critical to verifying the model's production relevance. Furthermore, we must determine whether the model successfully navigates the specific within-problem cases where energy and runtime disagree, confirming that the energy signal supplies ranking information a runtime proxy cannot.

\section{Methodology}
\label{sec:methods}
The approach combines three components built on a single deterministic measurement harness. We first construct a corpus of energy-labeled C++ programs, the Green Tea dataset (Section~\ref{sec:dataset-construction}), whose labels come from the Sniper and McPAT energy simulation harness (Section~\ref{sec:simulation}). To keep the comparison sound across stages, the same harness scores generations during the reinforcement-learning stage and again at evaluation. The model is then trained in two stages, an energy-aware supervised fine-tuning step on contrastive pairs (Section~\ref{sec:sft}) and a Group Relative Policy Optimization stage (Section~\ref{sec:grpo}) that closes the simulation-in-the-loop cycle. Because the reinforcement stage rewards the model with the measured cost of its own generations, it pushes the model to anticipate how the code it writes will behave under measurement rather than to reproduce fixed targets. We describe the evaluation protocol, metrics, and baselines in Section~\ref{sec:setup}.

\subsection{Green Tea Dataset Construction}
\label{sec:dataset-construction}
Energy-aware code optimization has no ready training dataset, since general-purpose code corpora carry no energy labels. Producing those labels is not a matter of simply running the code, because hardware readings cannot reproduce them reliably at the scale training needs. System-state variation introduces non-deterministic noise, and many programs run for well under a second, below the resolution of common power counters~\cite{hahnel2012rapl}. 
To create this labeled corpus (Phase~1 in Figure~\ref{fig:overview}), we build on 
$1{,}474$ problems from PIE~\cite{shypula2023pie}, a corpus of C++ competitive-programming submissions drawn from Project CodeNet~\cite{puri2021codenet}. A \emph{problem} is a single task with a fixed specification and a set of test inputs, and its solutions are independent programs that all implement that specification. 

\paragraph{From solutions to energy labels.}
A program's energy depends on its input~\cite{rajput2024enhancing,rajput2026energy}, so a single input would give a brittle, input-specific label. We therefore simulate each solution on every test input of its problem (a median of $102$ inputs per problem), drawn from Project CodeNet and supplemented with generated cases from AlphaCode~\cite{li2022alphacode}. Averaging over the full test set gives a stable per-solution energy and, at the same time, confirms the solution is correct on every input. The result is $3{,}507{,}435$ deterministic energy measurements, which required approximately $263{,}000$ CPU-hours to simulate. Releasing this pre-computed corpus allows future researchers to train energy-aware models without re-incurring this substantial compute cost. For each solution, the final energy label is the mean over its test inputs.

\paragraph{Per-problem energy scoring.}
A problem has many correct solutions that span a wide range of efficiency, so a useful training signal records not only which solution is more efficient but by how much. A fixed energy threshold cannot capture this, because absolute energy varies by orders of magnitude across problems and would label nearly every solution to a trivial problem as efficient and nearly every solution to a hard one as inefficient. We therefore score each solution by its rank within its own problem, following the per-task binned scoring used for performance-conditioned optimization~\cite{shypula2023pie}. For a given problem, we order its solutions by per-input-averaged energy and assign an integer score from $1$ to $10$ by decile, where $1$ marks the least efficient tenth and $10$ the most efficient. The score is relative to the other solutions of the same problem, which makes it comparable across problems of very different absolute cost. The $1$ to $10$ scale also gives a graded training signal, since conditioning the model on a target score lets it produce a range of optimization intensities, from constant-factor refactorings at a middling score to algorithmic refactorings at the top of the scale, rather than a single best-versus-worst contrast.

\paragraph{Pair construction and splitting.} A training pair couples a baseline solution of score $\leq 6$ with a more efficient target of score $\geq 7$ on the same problem, keeping only pairs with at least a $10\%$ energy reduction. The $6/7$ boundary places baselines in the lower six deciles and targets in the top four. This gives a wide energy gap per pair while keeping the baseline realistic rather than pathologically inefficient. The $10\%$ floor drops pairs whose energy gap is smaller than the measurement variation (a coefficient of variation of $1.36\%$ on our hardware), since such pairs would add noise to the training signal. Baselines range from naive brute-force solutions to partially optimized ones below the median, which gives diverse inefficiency patterns. The resulting pairs reduce energy by a mean of $61.3\%$ (median $67.2\%$). We split at the problem level ($80/10/10$) so that no held-out problem appears in training through any pair. This yields $9{,}927$ training pairs over $1{,}139$ problems, $1{,}301$ validation pairs over $143$ problems, and $1{,}227$ test pairs over $143$ held-out problems. Because the split is by problem rather than by pair, and each problem contributes a different number of pairs, these counts depart slightly from an exact division of the $12{,}455$ pairs. Scores are assigned within each problem, so every problem's solutions span the full $1$ to $10$ scale. The problem-level split, therefore, spreads all score levels across the three splits without separate stratification. The reinforcement-learning stage does not use pairs. It draws $15{,}853$ prompts from training-problem solutions that score $\leq 5$, the least efficient ones, which leave the most room for improvement, and for each prompt the policy generates its own candidate refactorings during training. These prompts come from the same $1{,}139$ training problems as the $9{,}927$ fine-tuning pairs but use different baseline programs, and no held-out test problem appears in either stage. 

\subsection{Sniper and McPAT Simulation Infrastructure}
\label{sec:simulation}

The energy labels for the Green Tea dataset (Section~\ref{sec:dataset-construction}) and the reward used in reinforcement learning (Section~\ref{sec:grpo}) both come from one deterministic simulation harness. We pair Sniper~\cite{carlson2014sniper}, a parallel interval simulator that models the timing of a CPU core and its memory hierarchy and emits per-component activity counts, with McPAT~\cite{dong2010mcpat}, an analytical power model that converts those activity counts into energy in joules. Section~\ref{sec:related-simulation} reviews both and their reported accuracy in more detail. Each run reports energy in joules together with cycles, instructions, instructions per cycle (IPC), and power, which the later stages use to form the reward.

The configuration models an AMD EPYC 9554P (Zen4) core at $1.5\,$GHz. Absolute frequency does not affect relative energy rankings, since every solution runs at identical clock settings and McPAT is linear in activity counts, so the ordering between solutions is invariant to a global calibration factor. Training and evaluation use the same harness, which keeps the optimization signal consistent across the two stages. Sniper and McPAT are established infrastructure for architectural energy and performance research, and the harness's fidelity here rests on the Sniper accuracy results in Section~\ref{sec:related-simulation} and our own empirical validation described below.

\paragraph{\textbf{Why simulation rather than actual measurement.}}

Training requires an energy value for every program the model generates, and at the volume training demands, that value must be both accurate and inexpensive to produce. Hardware measurements do not satisfy either requirement for this workload. The programs are brief, often executing in under a second, and at that duration, the run-to-run variance and counter-update artifacts of \texttt{perf} and RAPL exceed the energy differences the reward must resolve~\cite{hahnel2012rapl}. Measurement is therefore least reliable precisely when the reward must distinguish between two candidates of similar efficiency. Accuracy, however, is not the only binding constraint. On-policy reinforcement learning improves the model from its own generations, so the reward must execute every candidate the model proposes, as it is proposed during training, to obtain its energy, which makes physical measurement infeasible regardless of its precision. Each prompt yields $K$ candidates, and hardware must run them one at a time, in isolation on a dedicated machine, because background processes perturb the readings. The cost of a single training step, therefore, multiplies by $K$, which over a full training run becomes prohibitive. Simulation addresses both limitations. Its determinism guarantees that identical source and input configurations return identical energy, removing measurement variance and rendering every label and reward exactly reproducible. Because the simulator runs on commodity processors and parallelizes across them, dataset construction and the reward loop remain within a practical compute budget. Simulation further decouples the energy signal from any particular machine. A configuration file selects the modeled processor and retargets the harness to alternative designs, including microarchitectures that do not yet physically exist. This portability is a practical advantage rather than a convenience. A measured signal binds a model to the processors available when it was trained, forcing fresh measurements and re-training for each new hardware generation. In contrast, a simulated signal removes that dependence, letting the same model be optimized ahead of its deployment hardware. Furthermore, recent empirical work on LLM-generated code demonstrates that energy performance is highly correlated across different hardware platforms, suggesting that these optimizations are inherently portable~\cite{solovyeva2025ai}. For these reasons, simulation of sandboxed workloads is standard practice in energy and performance research.

\paragraph{Empirical validation of the simulation pipeline.}
To ensure this theoretical advantage holds in practice, we empirically verified both the \emph{determinism} and \emph{accuracy} of the simulation pipeline. For \textbf{determinism}, repeated executions of the pipeline on identical binaries across heterogeneous compute nodes in a shared high-performance cluster yielded energy variations of less than $0.1\%$, isolating the reward signal from hardware differences and background noise. For \textbf{accuracy}, a hardware spot check using the CodeGreen RAPL measurement framework~\cite{rajput2026codegreen,Rajput_CodeGreen_Towards_Improving_2026} confirms the simulator's rankings, avoiding contradiction on $93\%$ of resolvable program pairs and strictly agreeing on $80\%$. To ensure maximum physical precision, these hardware measurements adhered to established energy-profiling best practices~\cite{rajput2024enhancing}: executing warmup runs before measurement, sampling at a stable frequency, isolating the workload from background process noise, avoiding tail states, and averaging across multiple repeated executions. Because our dataset construction explicitly enforces a wide energy gap between baselines and targets (Section~\ref{sec:dataset-construction}), any remaining ranking noise on marginal cases is safely dwarfed by the large reductions we train on. Together, these empirical validations demonstrate that the simulation pipeline grounds our theoretical framework in practice, delivering a robust, rank-correct optimization signal that bypasses the noise limits of physical measurement.

\subsection{Energy-Aware Supervised Fine-Tuning}
\label{sec:sft}

In Phase~2A (Figure~\ref{fig:overview}), we perform supervised fine-tuning (SFT) of a base model on the Green Tea pairs with a score-conditioned prompt. A problem has many correct solutions of differing efficiency, so we tag each target with the decile score that records how close it is to the most efficient solution for that problem. The tag lets the model separate strong optimizations from weak ones instead of treating every improvement alike. This reward-conditioned generation idea is based on Decision Transformer~\cite{chen2021decision} and hindsight relabeling~\cite{zhang2023wisdom}, which PIE~\cite{shypula2023pie} adapts to program edits. We adapt it to energy by using the energy decile as the score.

\begin{lstlisting}[style=prompt,caption={Score-conditioned fine-tuning prompt template.},label={lst:sft-prompt}]
This is an energy inefficient program we want to optimize to score {target_score}/10.
### Program:
{baseline_code}

### Energy Optimized Version with score {target_score}/10:
\end{lstlisting}

During training, the \texttt{target\_score} field in the prompt (Listing~\ref{lst:sft-prompt}) is the actual decile score of the optimized solution, which for the Green Tea pairs spans scores $7$ through $10$, and the model is provided with the full target code appended to the prompt. At inference, we only provide the prompt up to the requested maximum score of $10$, and the model must generate the optimized target code itself.
The training objective is Dynamic Fine-Tuning (DFT)~\cite{dft}, which weights each token's loss by the model's predicted probability for that token. This down-weights the rare tokens that dominate standard cross-entropy and is reported to improve generalization on code-generation tasks. Formally, $\mathcal{L}_{\text{DFT}} = -\sum_{t} \text{sg}\!\big(p_\theta(y_t \mid y_{<t}, x)\big) \cdot \log p_\theta(y_t \mid y_{<t}, x)$, where $\text{sg}(\cdot)$ stops the gradient through the per-token weight and $p_\theta$ is the policy's predicted next-token probability. The weight is computed once per step on the same forward pass and does not contribute to the gradient, so DFT re-weights the standard cross-entropy rather than introducing a new objective. 
The contrastive structure of the pairs makes the training task explicit. Each pair shows an inefficient program, a requested score, and a more efficient program for the same problem, so the model learns to map an inefficiency \emph{signature}~\cite{mehditabar2026validated}, such as a nested $O(n^2)$ scan, a container chosen for the wrong access pattern, or an input-output call inside a hot loop, to a transformation at the requested score. Conditioning across scores $7$ to $10$ separates the energy-intensive refactorings at the upper end from the low-cost refactorings at the lower end. 

\subsection{GRPO with Simulation Reward}
\label{sec:grpo}

Fine-tuning (Section~\ref{sec:sft}) makes the model learn to imitate the given solutions and never runs the model's own generations, so it gives no feedback on whether those generations work and save energy. The reinforcement-learning stage (Phase~2B in Figure~\ref{fig:overview}) adds that execution feedback. Group Relative Policy Optimization (GRPO)~\cite{shao2024deepseekmath} initializes from the fine-tuned checkpoint and optimizes directly against the simulator-derived reward. For each prompt, the policy samples a group of $K$ refactoring candidates through vLLM~\cite{kwon2023vllm}, and each candidate is compiled, tested, and simulated through Sniper and McPAT. We set $K = 16$. This is lower than the $K = 64$ used by DeepSeekMath for text-only GRPO, because every rollout here carries a full simulation rather than a forward pass only. The value $16$ balances the within-group variance of the group-normalized \textit{advantage}~\cite{shao2024deepseekmath} against the per-step simulation budget, and we treat it as a budget-driven choice rather than a tuned hyperparameter. 

We use GRPO rather than a critic-based method because a learned critic is unreliable here. Proximal Policy Optimization (PPO)~\cite{schulman2017ppo} estimates a value baseline with a learned critic, but in our setting energy values span several orders of magnitude and the achievable energy depends on problem complexity, input size, and algorithmic class, so those value estimates are unreliable. GRPO removes the critic by normalizing rewards within each group of the $K = 16$ candidates for the same prompt. The candidates share the same problem and baseline energy, so their rewards are directly comparable without a learned baseline, and averaging the update over the group reduces variance. Algorithm~\ref{alg:grpo} summarizes one optimizer step. Each step samples $K$ rollouts from the current policy, scores every rollout through the compile-test-simulate harness, normalizes the scores within the group to form advantages, and updates the policy toward the higher-scoring rollouts under a Kullback-Leibler penalty that keeps it close to the supervised reference.

\begin{algorithm}[t]
\caption{GRPO with simulation reward (one optimizer step).}
\label{alg:grpo}
\begin{mdframed}[style=sagealg]
\begin{algorithmic}[1]
\REQUIRE Prompt $p$ (baseline code, target score), policy $\pi_\theta$, reference $\pi_\text{ref}$, group size $K$, KL coefficient $\beta$
\STATE Sample $K$ rollouts $\{y_i\}_{i=1}^K \sim \pi_\theta(\cdot \mid p)$ via vLLM
\FOR{$i = 1, \ldots, K$ \textbf{in parallel}}
  \STATE Compile, test, and Sniper/McPAT-simulate $y_i$
  \STATE Compute $r_i = r_\text{correct}(y_i) + r_\text{energy}(y_i)$
\ENDFOR
\STATE Group-normalize: $\hat{A}_i = (r_i - \mathrm{mean}(r))/\mathrm{std}(r)$
\STATE Update $\theta$ by ascending $\sum_i \hat{A}_i \log \pi_\theta(y_i \mid p) - \beta\, \mathrm{KL}(\pi_\theta \| \pi_\text{ref})$
\end{algorithmic}
\end{mdframed}
\end{algorithm}

\paragraph{Reward design.}
The reward combines a correctness term and an energy term, $R = r_{\text{correct}} + r_{\text{energy}}$. The correctness term is piecewise in the test pass fraction $f \in [0,1]$,
\begin{equation*}
r_{\text{correct}}(f) =
\begin{cases}
\rho_{\text{fail}}, & \text{compile failure, runtime error, or ghost execution},\\
\rho_0 + \lambda\,f, & f \in [0,1),\\
\rho_{\text{pass}}, & f = 1,\\
0, & \text{simulation timeout}.
\end{cases}
\end{equation*}

Without strict correctness gating, an energy reward invites reward hacking. A program that aborts immediately performs no work and consumes near-zero energy. If rewarded purely for this reduction, the policy learns to generate trivially broken programs rather than genuine optimizations. We guard against this by detecting \emph{ghost executions}, which are Sniper traces with zero retired instructions or zero energy that indicate the program exited before reaching any measured region, commonly due to a build artifact or a degenerate input handler, and assigning them the hard-failure reward $\rho_{\text{fail}}$. Following established reinforcement learning practice, what matters for training is not the absolute magnitude of the rewards, but the relative ordering and the strategic gaps between outcome categories, because GRPO normalizes rewards within each group before the update. Consequently, we define four core constants to structure these gaps: $\rho_{\text{fail}} = -1.0$, $\rho_0 = -0.8$, $\lambda = 1.1$, and $\rho_{\text{pass}} = +0.5$. Three properties fix this structure. First, the partial ramp begins above the failure floor ($\rho_0 - \rho_{\text{fail}} = 0.2$), which preserves a useful gradient between a program that compiles but is partially correct and one that does not compile at all. Second, the slope $\lambda = 1.1$ caps the partial branch just under $+0.3$, leaving a deliberate penalty gap of $0.2$ to the full-correctness bonus, preventing the policy from settling for near-complete correctness. Third, $\rho_{\text{pass}}$ is strictly positive, ensuring that a fully correct solution with zero energy improvement still earns a positive reward ($0.5 + \tanh(0) = 0.5$), thereby maintaining the correctness incentive even when no energy gain is found. We establish these constants analytically by this reasoning rather than by an empirical sweep (Section~\ref{sec:threats}).

The energy term activates only when $r_{\text{correct}} = \rho_{\text{pass}}$, so partially correct solutions never receive an energy reward. Our framework accommodates any simulation-derived metric that influences efficiency, such as raw energy, runtime, or the energy-delay product (EDP), where $\text{EDP} = E \times t$, with runtime $t$ derived from Sniper's cycle count at the configured clock frequency. For example, using EDP as the optimization proxy, with baseline $E_b, t_b$ and generated candidate $E_g, t_g$, the performance gain $g$ and resulting energy reward are
\begin{equation*}
g = \frac{E_b\,t_b - E_g\,t_g}{E_b\,t_b},
\qquad r_{\text{energy}} = \tanh(g).
\end{equation*}
The $\tanh$ function bounds the energy reward to $(-1, 1)$, ensuring that extreme performance ratios cannot dominate the gradient. The correctness and energy terms combine into a per-rollout reward in $[-1.0, 1.5)$, reaching about $1.26$ for a fully correct output at a gain of $g = 1.0$.

\paragraph{Choice of reward signal.}
As shown in Phase~2B of Figure~\ref{fig:overview}, the reinforcement learning loop optimizes an \emph{Energy-Aware Reward}. The framework accommodates various simulation-derived metrics, including raw energy, runtime, or their product (evaluated further in Section~\ref{sec:rq3}). We detail the energy-delay product (EDP) throughout this section to illustrate the reward mechanics, as it naturally aligns with deployment contexts that seek energy reductions without severe runtime regressions. EDP credits a refactoring only when it lowers energy without trading the saving back as a longer runtime. It also offers two structural properties that make it convenient for training. First, $\text{EDP} \propto E \times \text{cycles}$ amplifies the reward at a fixed frequency (since a $2\times$ improvement in both energy and cycles yields a $4\times$ ratio), leaving fewer rollouts with near-zero rewards that contribute nothing to learning. Second, by combining time and power, it captures within-problem variation in the instruction mix that runtime alone cannot rank. Conversely, we do not reward instructions per cycle (IPC) directly, because higher throughput does not imply lower energy and the two can move in opposite directions within a problem.

\paragraph{Training, checkpointing, and compute.}
Both stages train Low-Rank Adaptation (LoRA)~\cite{hu2022lora} adapters on every attention and feed-forward projection ($q, k, v, o, \text{gate}, \text{up}, \text{down}$), which freezes the base weights and leaves about $2.8\%$ of the model trainable at rank $64$ with scaling $128$. We do not sweep the rank, since the goal is to evaluate the training procedure rather than to optimize adapter capacity. 
Both stages adopt standard optimizer settings from their original sources rather than tuning them here. Fine-tuning uses AdamW~\cite{loshchilov2019adamw} at learning rate $3 \times 10^{-5}$ with cosine decay and weight decay over three epochs, at an effective batch size of $32$ from a per-device batch of $4$ and gradient accumulation over $8$ steps. GRPO also uses AdamW, at policy learning rate $10^{-6}$ with a Kullback-Leibler (KL) penalty at coefficient $\beta = 0.04$, following the GRPO defaults of DeepSeekMath~\cite{shao2024deepseekmath}, and runs for one epoch over the $15{,}853$ prompts. Training uses a single H100 with $48$ CPU cores for parallel simulation, and we report the final-epoch checkpoint. Across the full study, spanning nine GRPO configurations, a seven-model supervised fine-tuning sweep, and evaluation, training and inference consume approximately $1{,}800$ H100-GPU-hours; the main pipeline alone accounts for $168$ of these over a seven-day GRPO run plus a further $33.3$ for its fine-tuning. Dataset construction, the $3{,}507{,}435$ simulations, requires roughly $263{,}000$ CPU-hours over about four months.

\subsection{Evaluation Setup}
\label{sec:setup}

\paragraph{Protocol.}
In the final evaluation (Phase~3 in Figure~\ref{fig:overview}), we evaluate each model at the generation settings, such as temperature, recommended in its documentation. Each generated program is compiled with \texttt{g++ -O3 -std=c++17 -static} ($120\,$s timeout), tested against all inputs for its problem (a median of $102$ inputs), and energy-measured by Sniper simulation ($120\,$s timeout per input). This two-minute simulation timeout follows established protocols for competitive-programming evaluation~\cite{shypula2023pie}, safely accommodating the vast majority of valid solutions while preventing infinite loops from exhausting the compute budget. The uniform \texttt{-O3} flag, which applies the highest level of compiler optimization, ensures that observed energy differences reflect \textbf{source-level choices} rather than missed compiler optimizations, and \texttt{-static} produces self-contained binaries that the simulator runs without external library dependencies. The test set comprises $1{,}227$ samples over $143$ held-out problems, and each sample receives a single generation at evaluation. This is the per-invocation deployment behavior we measure, in contrast to GRPO training, where the policy samples $16$ candidates per prompt for the group-relative update. We do not select among multiple generations at evaluation. Because the compile-test-simulate procedure is deterministic and needs only a standard x86 machine with Sniper and McPAT, the evaluation is reproducible without specialized hardware.

\paragraph{Metrics.}
\label{sec:metrics}
We report three metrics, together with a diagnostic decomposition of the primary one. The \emph{Energy Reduction Rate} (ERR) is a standard relative-improvement measure, the energy counterpart of the runtime gains reported in performance optimization~\cite{shypula2023pie}. For a valid output, $\text{ERR} = (E_{\text{base}} - E_{\text{gen}})/E_{\text{base}}$ is the fraction of energy the generated program saves relative to its baseline, and it is defined only on outputs that compile and run. ERR credits efficiency but ignores outputs that break, so a model that fails often can still post a high ERR. We therefore introduce the \emph{Correctness-Adjusted Reduction in Energy Total} (CARET), our primary measure, which folds correctness into the energy reduction so that an output is credited for its energy saving only to the extent it stays correct, the property a deployable optimizer needs and that efficiency measures gated on binary correctness do not provide. CARET extends ERR over all $N$ outputs, including failures, by weighting each output's saving by the fraction of tests it passes, with compilation failures contributing zero,
\begin{equation}
\label{eq:caerr-def}
\text{CARET} = \frac{1}{N} \sum_{i=1}^{N} \text{ERR}_i \cdot \frac{t_i^{\text{pass}}}{t_i^{\text{total}}},
\qquad \text{ERR}_i = 0 \text{ for non-compiling outputs}.
\end{equation}
A solution at $50\%$ ERR that passes half its tests contributes $25\%$, so CARET is the average energy reduction per model invocation, with failures counted as zero rather than excluded.
Pass@k, a standard code-generation metric, treats correctness as binary, ignores energy, and reports how often a model produces correct code given several attempts. CARET answers a different question. It weights each sample by both its test pass fraction and its energy reduction, and reports how much energy a single deployment call is expected to save. Prior work that reports reduction rates on valid outputs alone, for example PIE~\cite{shypula2023pie}, excludes failed compilations from the denominator, which inflates the apparent gain of a model that fails often. We therefore report the compile rate and an input-weighted test pass rate, the share of test inputs passed across all outputs, alongside CARET so that a reader can reconstruct it, and we recommend the same for future energy reduction reports. This input-weighted rate differs from the conditional pass rate $P(\text{pass} \mid \text{compile})$ in the decomposition below (Equation~\ref{eq:caerr-decomp}), which conditions on compilation and weights every problem equally.
Finally, \emph{Beat-GT} (beating the ground-truth reference) is the fraction of valid outputs whose energy efficiency matches or exceeds the human-expert reference solution for the same problem. An output counts as beating the reference when its simulated energy is at most the reference's ($E_{\text{model}} \le E_{\text{ref}}$), with ties credited to the model. The reference is the runtime-optimized human solution from PIE~\cite{shypula2023pie}, so Beat-GT measures whether energy-aware training reaches or exceeds a runtime-tuned human ceiling on the energy axis.

To attribute improvements across capability dimensions, we decompose CARET into three factors,
\begin{equation}
\label{eq:caerr-decomp}
\text{CARET} \approx \underbrace{P(\text{compile})}_{\text{compile rate}} \times \underbrace{P(\text{pass} \mid \text{compile})}_{\text{conditional pass rate}} \times \underbrace{\mathbb{E}[\text{ERR} \mid \text{correct}]}_{\text{ERR on valid}}.
\end{equation}
This product is an approximation, not an identity. The three factors are correlated, since an output that compiles is also more likely to pass its tests and to save energy, so the product of the marginal factors departs from the joint quantity that CARET measures. We therefore read the decomposition only as a directional guide to which factor moves, never as a substitute for the directly computed CARET, and we always report the three factors alongside the aggregate (Section~\ref{sec:caerr-detail}).

A within-problem energy reduction separates further into a cycle-driven and a power-driven part. Writing $r_{\text{cycles}} = C_{\text{gen}}/C_{\text{base}}$ for the ratio of generated to baseline cycles and $r_{\text{power}} = (E_{\text{gen}}/E_{\text{base}})/r_{\text{cycles}}$ for the residual power ratio, the reduction factors as $\text{ERR} = (1 - r_{\text{cycles}}) + r_{\text{cycles}}(1 - r_{\text{power}})$. The first term is the reduction from running fewer cycles, and the second is the reduction from drawing less power at a fixed cycle count, which separates a runtime-visible saving from a runtime-invisible one.

\paragraph{Baselines.}
\label{sec:baselines}
We evaluate our approach on a representative rather than exhaustive sample of state-of-the-art code models, guided by compute constraints and benchmark rankings. Specifically, we select models that rank among the top code models at the time of writing and that can be fine-tuned within a single-H100 budget. The set is mostly open-weight, as open weights are required for local fine-tuning. These include the Qwen2.5-Coder family ($0.5$B, $7$B, $14$B, and $32$B), DeepSeek-Coder-6.7B, DeepSeek-Coder-V2-Lite, and Llama-3.1-8B. To ensure our findings are not tied to open weights alone, we also include popular models reached via managed fine-tuning services: Gemma-3-12B-IT and Gemini-2.5-Flash, with the latter serving as a closed-weight state-of-the-art reference. Configurations are named by model, parameter count, and variant (e.g., Qwen14B-Energy-SFT), with the exact prompt text in the replication package.

We distinguish two model variants: \emph{base} models pre-trained for next-token prediction, and \emph{instruction-tuned} models post-trained for chat. While instruction tuning improves zero-shot correctness, it locks the model into a response format that our completion-style training later works against.

To isolate the source of any energy gains, each fine-tuned model is compared against four baselines under an identical compile-test-simulate protocol. The \emph{zero-shot} baseline provides a plain optimization instruction, while \emph{green-prompting}~\cite{cursaru2024controlled,vartziotis2024learn} adds an explicit energy-efficiency directive to test if prompting alone suffices. \emph{Runtime-SFT} fine-tunes the model on the same dataset ranked by runtime rather than energy, reproducing the PIE~\cite{shypula2023pie} methodology within our protocol. In the reinforcement learning stage, our \emph{R-SFT + runtime} evaluation (Section~\ref{sec:rq3}) extends this to reproduce the runtime-focused GRPO methodology of Afterburner~\cite{du2025afterburner}. Finally, the \emph{instruction-tuned fine-tuning} baseline applies our training data to the instruction-tuned variant, separating instruction-following capability from energy-specific learning. The two largest models (Qwen2.5-Coder-32B and DeepSeek-Coder-V2-Lite) are evaluated only as zero-shot and green-prompted external reference points.

\section{Results}
\label{sec:results}
\subsection{Answering RQ1: Do Standard Throughput Proxies Reliably Rank Energy Efficiency?}
\label{sec:rq1}

A single programming problem typically admits numerous correct implementations that achieve the same result through different computational pathways, leading to vast differences in energy consumption. To rank these solutions effectively, we must first decompose how this energy is spent. A program's energy is what its compiled executable binary draws while it runs. The binary issues a stream of machine instructions, and the processor completes them over clock cycles. The work a solution performs is set by how many instructions it runs, and the time it takes is set by how many cycles it spends, which at a fixed clock frequency is proportional to runtime. Energy is the power drawn over that time, $E = P \times t$, and the power depends on how hard each cycle drives the processor, which follows from the mix of instructions it executes. A solution can therefore lower its energy in two distinct ways, by doing less work so it spends fewer cycles, or by running a lower-power instruction mix at the same cycle count.

This breakdown shows that a program's energy is built from, and shaped by, several measurable quantities, each of which could rank competing solutions. These are the instruction count, the cycle count, the runtime, the instructions completed per cycle (IPC), and the power. We use the term \emph{signal} for any such measure once it serves to rank solutions or to form a training reward. RQ1 asks which of these signals ranks competing solutions as their measured energy does, because that signal sets the target a model can be trained to optimize. A signal that ranks solutions incorrectly would direct training toward the wrong target. We study this on the Green Tea corpus (Section~\ref{sec:dataset-construction}) of $1{,}474$ problems, with the within-problem comparisons computed over the $1{,}444$ problems that retain a solution above the $190\,$W power floor (Table~\ref{tab:landscape}). The simulator is deterministic, so any disagreement between two signals on the same programs reflects the programs themselves rather than measurement noise.

\paragraph{The within-problem energy spread.}
Competing solutions to one problem differ widely in energy, which is what makes the choice of ranking signal consequential. The most and least efficient human solutions to the same problem differ in energy by a median of $6.52\times$ and a mean of $18.10\times$ (Table~\ref{tab:landscape}), a heavy-tailed distribution in which order-of-magnitude differences are common. The room to improve is uneven across problems, and we size it by the within-problem \emph{oracle reduction}, the energy a problem's most efficient solution saves relative to its least efficient one. On $39.2\%$ of problems this oracle reduction exceeds $90\%$, on $31.6\%$ it falls in the $50$ to $90\%$ range, and on the remaining $29.2\%$ it stays below $50\%$. These reductions are reachable through changes that range from algorithmic improvements, such as replacing an $O(n^2)$ method with an $O(n \log n)$ one, to constant-factor work such as better cache locality on the same algorithm~\cite{bentley1982writing,leiserson2020there,shypula2023pie}.

\begin{table*}[t]
\centering
\caption{Energy variation among valid solutions within the same problem. CI represents the 95\% confidence interval.}
\label{tab:landscape}
\begin{tabular}{lcc}
\toprule
Metric & Median & Mean \\
\midrule
Energy ratio (worst/best) & $6.52\times$ (95\% CI [5.73, 7.14]) & $18.10\times$ \\
Energy difference (worst $-$ best) & 0.211\,J & 0.622\,J \\
Per-problem power range & 3.65\,W & 5.64\,W \\
\midrule
Oracle energy reduction (worst $\to$ best) & 84.66\% (CI [82.6, 86.0]) & 70.76\% \\
\midrule
\multicolumn{3}{l}{$> 90\%$ energy reducible: 39.2\% of problems \quad $> 50\%$: 70.8\%} \\
\bottomrule
\end{tabular}
\end{table*}

\paragraph{Across problems, runtime tracks energy.}
The first question is whether an easier-to-measure signal could serve as a proxy for energy. Across problems of different difficulty, the candidate signals agree, because a harder problem runs more instructions, spends more cycles, takes longer, and draws more energy. These quantities span several orders of magnitude across the corpus, and a log-log regression of energy on runtime is almost perfectly linear ($R^2 = 99.98\%$, $r = 0.9999$). This agreement is why prior work on learned code optimization optimizes runtime and treats it as a proxy for energy~\cite{shypula2023pie,rajput2026energy}. Training, however, operates within a single problem, comparing solutions that implement the same task and spend similar numbers of cycles, and there the cross-problem agreement breaks down, so a cheap proxy can rank two solutions the opposite way to energy.

\begin{figure}[t]
\centering
\includegraphics[width=0.45\columnwidth]{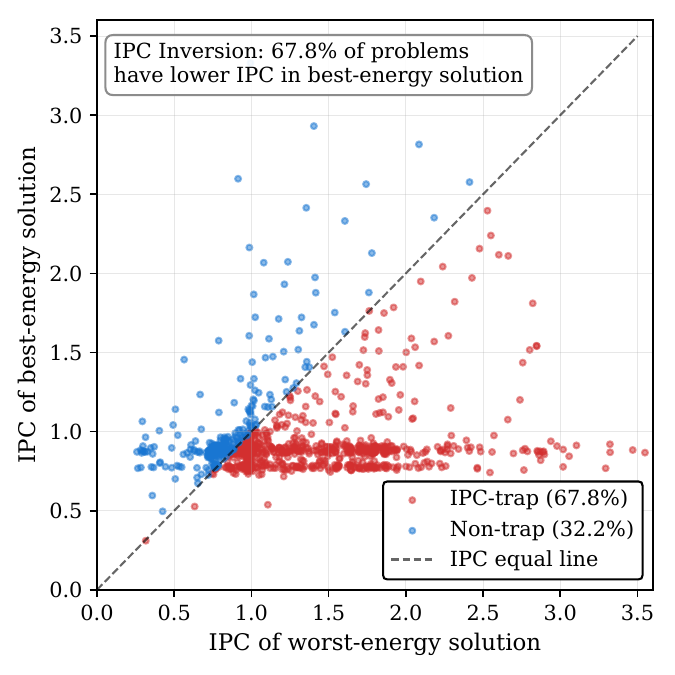}
\caption{IPC vs.\ energy for the best- and worst-energy solutions per problem. The IPC-trap region highlights problems where the most energy-efficient solution has \emph{lower} IPC, showing that throughput proxies invert true ranking.}
\label{fig:ipc-trap-scatter}
\end{figure}

\paragraph{The IPC trap.}
Consider first instructions per cycle (IPC), a seemingly natural throughput signal. IPC measures how many instructions a core retires per cycle, so a throughput-oriented reading treats a higher IPC as the more efficient solution, since the execution units stay busier. However, this assumption fails for energy. A higher IPC provides no information about the total instruction count, and for these workloads, energy consumption is driven far more by the total volume of executed instructions than by their per-cycle throughput~\cite{horowitz20141,hennessy2011computer}. Our empirical data directly contradict this throughput-centric intuition. On $67.8\%$ of problems (95\% confidence interval $[65.3, 70.2]\%$) the most energy-efficient solution has a \emph{lower} IPC than the least efficient one, a mean of $0.928$ against $1.167$, a pattern we call the \emph{IPC trap} (Figure~\ref{fig:ipc-trap-scatter}). The underlying mechanism is a drastic reduction in instruction count. A more efficient solution frequently replaces compute-heavy loops with efficient data-structure lookups, such as substituting a linear scan with a hash-map probe, which simultaneously lowers both the instruction count and the IPC. The energy saved by executing fewer instructions far outweighs the penalty of lower per-cycle throughput. Since IPC exhibits only a weak Pearson correlation ($r = 0.205$) with power across competing solutions, it is unsuitable as a ranking signal; training a model to maximize IPC would actively steer it toward compute-dense code, worsening energy efficiency on most problems. Conversely, while simply minimizing the total instruction count avoids the IPC trap, it remains an incomplete proxy because it treats all instructions as drawing equal power, ignoring the microarchitectural variations that drive actual power consumption.

\begin{figure}[ht]
\centering
\includegraphics[width=0.9\columnwidth]{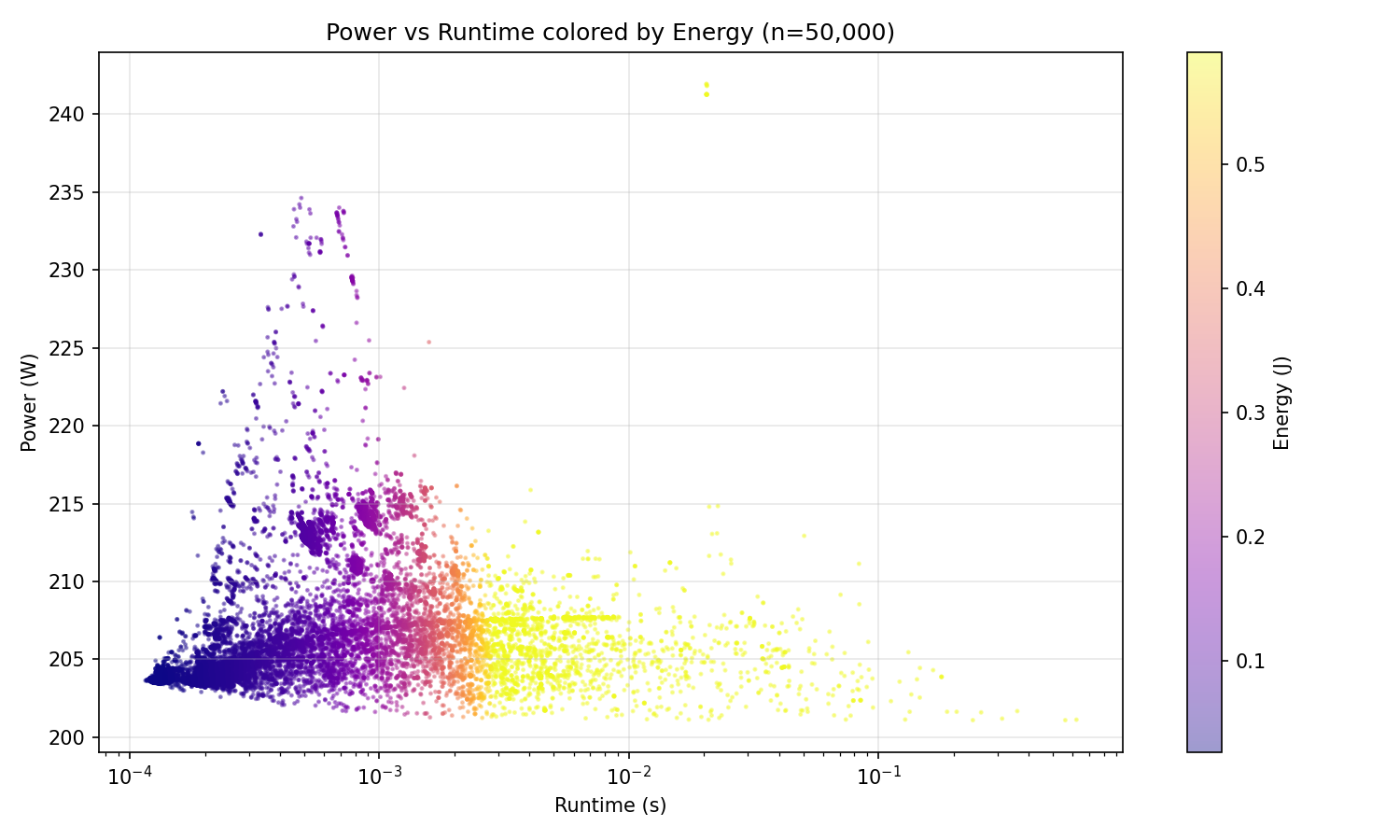}
\caption{Power draw vs.\ runtime, colored by energy. The vertical spread at any fixed runtime demonstrates power variations invisible to a purely runtime-based reward.}
\label{fig:power-runtime}
\end{figure}

\paragraph{Runtime is blind to power.}
A faster solution is not always the more energy-efficient one. Runtime is a safer signal than IPC, yet it still fails within a problem. When two solutions spend almost the same number of cycles, within $1\%$ of each other, runtime cannot separate them at all, yet $41.2\%$ of these cycle-matched pairs differ in energy, by more than $0.5\%$ and by as much as $11.1\%$, a difference that comes entirely from power draw (Figure~\ref{fig:power-runtime}). Runtime is blind to this gap by construction, since it reads only the cycle count, and at $41.2\%$ of cycle-matched pairs the gap is common rather than a corner case. A runtime ranking also inverts the energy order outright on $9{,}380$ within-problem pairs, which would reverse the ranking of the corresponding energy-ranked fine-tuning pairs and hand the model a higher-energy solution as its training target, a bias that energy-based selection avoids. These metrics represent a lower bound rather than the typical case. Because the simulator is deterministic, it reports power with almost no measurement noise (a coefficient of variation of $1.60\%$), so these counts reflect genuine differences between solutions rather than measurement artifacts. On real hardware power varies more across runs~\cite{hahnel2012rapl}, so a runtime signal would misrank at least as often there.

 These fractions may seem small on a single short program but grow at deployment scale. The power variation behind the within-problem energy differences increases on longer-running and memory-bound workloads, where the constant-power assumption of a runtime signal is weakest, and such programs run repeatedly and at volume, so a saving that is negligible on one invocation accumulates across a deployed system.

\paragraph{Decomposing the within-problem difference.}
To see what a correct signal must capture, we split each within-problem energy difference into a cycle component and a power component. A Laspeyres index decomposition~\cite{ang2004decomposition, ang2007energy}, a standard method for splitting a change in a product into the contributions of its factors, separates the energy saved by running fewer cycles from the energy saved by drawing less power, holding power at its baseline level to isolate the cycle part. On $87.7\%$ of problems it attributes a median $98.7\%$ of the energy difference to cycle reduction, where runtime and energy agree on the order. On the remaining $12.3\%$ of problems power contributes more than $5\%$ of the difference, and there cache behavior and instruction mix decide the winner while runtime points the wrong way. Power varies even at a fixed runtime, spreading by a median of $7\,$W and up to $40\,$W (Figure~\ref{fig:power-runtime}). This spread tracks the program's instruction mix rather than measurement noise. On a single program the per-input energy varies by a coefficient of variation of only $1.36\%$ to $1.60\%$, a consequence of simulator determinism, whereas across different programs at the same runtime it rises to $6.72\%$, so what a program does, rather than which input it is given, drives the variation~\cite{rajput2026energy}. Because energy is the product of power and time, it intrinsically captures the ranking information contained in both dimensions. In contrast, a cycle or runtime signal captures only the time dimension, blinding it to power variations. Consequently, a direct energy metric correctly ranks these architectural tradeoffs where runtime or throughput proxies fail.

Two conclusions follow for training any model to produce energy-efficient code. First, energy must be the optimization signal itself rather than a value inferred from runtime or a throughput measure, because runtime misranks the power-driven $12.3\%$ of problems and IPC inverts the order on most. Such a signal can be produced at the volume training needs through deterministic simulation, both as rank-correct contrastive pairs and as an online reward. Second, energy rather than runtime is the durable target. A runtime-trained model inherently assumes constant power. While this assumption holds for cycle-dominated code, it already breaks on the $12.3\%$ of power-driven problems in our dataset. This blind spot widens significantly on longer-running or memory-bound workloads, as well as on mobile, heterogeneous, and GPU systems where power varies drastically across operations. Together, these findings establish that an effective training procedure requires a direct energy signal, one that neither runtime nor throughput metrics can safely replace.

\begin{findingbox}[RQ1 answer]
Runtime and energy rank solutions the same way across problems but not within a problem, which is where training operates. Within a problem a runtime signal is blind to the power differences that decide the ranking, and a throughput measure is worse still, with IPC inverting the energy order on $67.8\%$ of problems. Energy must therefore be the optimization signal itself, since it captures both the time a solution takes and the power it draws while a runtime or throughput measure sees only one. Deterministic simulation supplies this signal at the volume model training needs.
\end{findingbox}
\subsection{Answering RQ2: Can Supervised Fine-Tuning Alone Produce a Deployable Energy Optimizer?}
\label{sec:rq2}

RQ1 shows that energy and runtime can rank competing solutions for the same problem differently, so a model trained to generate energy-efficient code needs an objective that reflects both the time a solution takes and the power it draws. This section asks whether supervised fine-tuning on energy-ranked pairs teaches a model to optimize for that objective, and whether fine-tuning alone is enough to deploy. A deployable model must reduce energy without degrading correctness, so we report CARET (defined in Section~\ref{sec:setup}) alongside the compile rate and the rate at which an output beats the human-written reference. We evaluate this training procedure across seven different models, comparing each against its own zero-shot and prompting baselines (Section~\ref{sec:baselines}). This setup mirrors a realistic deployment scenario where a developer seeks to improve the energy efficiency of a model that was already selected due to other operational constraints, such as licensing, latency, or benchmark performance.

\paragraph{Fine-tuning targets the base variant.} Our fine-tuning procedure targets the base variant rather than the instruction-tuned one, because instruction tuning locks a model into a chat-style response format that the completion-style score-conditioned target works against, a conflict that typically surfaces as catastrophic forgetting when the two are combined~\cite{mccloskey1989catastrophic,zhai2023investigating}. To verify this empirically before applying the procedure across all models, we evaluate it on Qwen2.5-Coder-14B, the one model in our set with base, instruction-tuned, and fine-tuned variants all available. The evidence in Table~\ref{tab:main-results} confirms this design choice. The base model is non-functional out of the box, compiling $1.5\%$ of its outputs under a plain instruction and $0.1\%$ under the energy-framed prompt. The instruct model compiles far more reliably at $78.9\%$, but its correct outputs increase energy on average ($-2.92\%$ CARET, driven by a thin tail of large regressions), and an explicit energy instruction makes it worse ($-4.07\%$ CARET), so general ability and prompting do not produce energy optimization on their own. Fine-tuning the base model on the energy-ranked pairs is the first configuration whose outputs reduce energy on average, reaching $4.45\%$ CARET from the harder non-functional starting point, whereas fine-tuning the instruct variant on the same data collapses it, dropping compilation to $38.0\%$ and CARET to $0.19\%$. Conditional on producing valid code the two starting points learn the optimization pattern about equally ($6.39\%$ versus $7.95\%$ mean reduction), so the collapse is a loss of code-generation ability rather than of optimization. We therefore fine-tune the base variant throughout.

\paragraph{Energy optimization transfers across models.} Table~\ref{tab:cross-family} reports the change from each model's zero-shot baseline to its fine-tuned variant. Off-the-shelf prompting does not produce energy optimization on most models. The zero-shot CARET ranges from a $-2.92\%$ regression on Qwen2.5-Coder-14B to reductions near zero, the one exception being the already-strong Gemma-3-12B-IT, whose instruct baseline reaches $6.44\%$. Energy optimization therefore has to be learned from energy-contrastive data rather than prompted for, and fine-tuning on energy-ranked pairs, which we call Energy-SFT, supplies that pattern and transfers across models. CARET rises in six of the seven, including the sign flip on Qwen2.5-Coder-14B, a rise from a flat zero-shot baseline on Llama-3.1-8B, and a further gain on the already-strong Gemma-3-12B-IT, from $6.44\%$ to $9.08\%$. The one exception is DeepSeek-Coder-6.7B, where fine-tuning lifts CARET only marginally, from $1.08\%$ to $1.29\%$, while collapsing the compile rate from $95.4\%$ to $8.0\%$, so it trades nearly all of the model's correctness for no optimization gain and leaves the model unusable. The likely cause is catastrophic forgetting, where adapting a heavily post-trained model to the score-conditioned completion format overwrites the behavior that let it compile~\cite{mccloskey1989catastrophic,zhai2023investigating}. A general decline in compile rates under fine-tuning is an expected characteristic of score-conditioned generation. By prompting the model to aggressively target the maximum possible optimization score, it occasionally attempts complex transformations that break syntactic viability on certain problems. Prior work observes this exact same tradeoff for performance-conditioned generation, confirming it as a natural artifact of the target formulation rather than a failure to learn~\cite{shypula2023pie}. Because Gemma and Gemini are fine-tuned via a restricted third-party API service, we lack control over hyperparameters like adapter rank and loss masking. We therefore interpret their results as directional evidence of transferability rather than strictly controlled replications, with the closed-weight Gemini-2.5-Flash serving as an additional reference point.

Simply scaling up a model's parameters does not naturally close the performance gap that energy-contrastive fine-tuning addresses, meaning scale alone cannot substitute for this explicit training signal. To verify this, we evaluate two larger off-the-shelf models, Qwen2.5-Coder-32B and DeepSeek-Coder-V2-Lite. Both achieve zero-shot CARET scores of only $-16.19\%$ and $4.27\%$, respectively, beating the human reference on fewer than one in ten outputs. These results trail the fine-tuned $14$B model, confirming that energy optimization must be explicitly learned.

\begin{table}[t]
\centering
\small
\caption{Cross-model transfer of Energy-SFT. We report the absolute percentage point (pp) change from each model's zero-shot baseline. Bold indicates a positive CARET gain.}
\label{tab:cross-family}
\begin{tabular}{llccl}
\toprule
Family & Baseline & CARET\% (base\,$\to$\,SFT) & Compile\% (base\,$\to$\,SFT) & Effect \\
\midrule
Qwen2.5-Coder-14B   & Instr-ZS & $-2.92 \to \mathbf{4.45}$ & $78.9 \to 60.3$ & turns positive \\
Llama-3.1-8B        & Instr-ZS & $0.00 \to \mathbf{5.97}$  & $78.7 \to 48.2$ & from zero \\
Qwen2.5-Coder-7B    & Base-ZS  & $3.18 \to \mathbf{8.71}$  & $90.0 \to 64.0$ & $2.7\times$ \\
Qwen2.5-Coder-0.5B  & Base-ZS  & $2.91 \to \mathbf{4.07}$  & $73.4 \to 39.3$ & $1.4\times$ \\
Gemini-2.5-Flash    & Instr-ZS & $0.37 \to \mathbf{0.79}$  & $7.6 \to 5.0$   & $2.1\times$  \\
DeepSeek-Coder-6.7B & Base-ZS  & $1.08 \to \mathbf{1.29}$  & $95.4 \to 8.0$  & gain but compile collapse \\
Gemma-3-12B-IT      & Instr-ZS & $6.44 \to \mathbf{9.08}$  & $84.4 \to 72.9$ & $1.4\times$ \\
\bottomrule
\end{tabular}
\end{table}

\paragraph{Energy ranking versus runtime ranking.} While cross-model transfer demonstrates that energy-contrastive fine-tuning is effective, it leaves open whether the explicit energy ranking is necessary, or if ranking by runtime would perform just as well. Table~\ref{tab:energy-vs-runtime-panel} compares the two ranking strategies across the four mid-size models and the smaller Qwen2.5-Coder-0.5B baseline model. The runtime-ranked variant (Runtime-SFT) reproduces the PIE baseline~\cite{shypula2023pie}, which trains models to optimize for performance. Because faster execution generally reduces energy consumption, this runtime-ranked approach serves as a strong conventional proxy for energy optimization.

In terms of aggregate CARET, our Energy-SFT approach matches or exceeds this runtime-ranked baseline on three of the four mid-size models: it outperforms on Qwen2.5-Coder-7B and DeepSeek-Coder-6.7B, slightly outperforms on Llama-3.1-8B, and trails only on Qwen2.5-Coder-14B. The per-problem paired tests offer a more nuanced view. Energy ranking is favored on Qwen2.5-Coder-7B (paired Wilcoxon $p{=}0.029$) and runtime ranking on Qwen2.5-Coder-14B ($p{=}0.004$), while Llama-3.1-8B shows no detectable per-problem difference ($p{=}0.97$, $n{=}143$). DeepSeek-Coder-6.7B trends toward energy on its paired problems. Crucially, beyond raw energy savings, energy ranking holds a distinct and compelling advantage in deployment reliability: it consistently achieves higher correctness, better deployment safety, and lower rates of catastrophic regression across every open-weight model we test. Thus, energy ranking not only matches or exceeds the runtime proxy's energy reductions on most models, but it also serves as the fundamentally safer choice for modifying working code in production.

The closed-weight Gemini-2.5-Flash model provides an additional comparison point. Tuned via the same restricted third-party API service, its energy-ranked variant suffers a compile rate collapse (to $5.0\%$), whereas the runtime-ranked variant retains a $77.2\%$ compile rate and achieves $16.74\%$ CARET. However, on the $33$ problems where both variants produce valid outputs, there is no detectable difference in their per-problem energy reduction ($p{=}0.46$). This reveals that wherever the model successfully emits valid code, the explicit energy signal optimizes just as powerfully as the runtime signal. The apparent gap in aggregate CARET is strictly a byproduct of differing compile rates on this specific API service, rather than any deficiency in the energy optimization signal itself.

\begin{table}[t]
\centering
\small
\caption{Energy-SFT vs.\ Runtime-SFT across mid-size models and the smaller Qwen-0.5B baseline model. Bold marks the higher CARET.}
\label{tab:energy-vs-runtime-panel}
\begin{tabular}{lrrcl}
\toprule
Family & E-SFT \% & R-SFT (PIE) \% & $\Delta$ (pp) & Better-tuned with \\
\midrule
Qwen2.5-Coder-7B    & \textbf{8.71}  & 7.92  & $-0.79$ & Energy-ranked \\
Qwen2.5-Coder-14B   & 4.45           & \textbf{7.64} & $+3.19$ & Runtime-ranked \\
DeepSeek-Coder-6.7B & \textbf{1.29} & 0.65 & $-0.64$ & Energy-ranked \\
Llama-3.1-8B        & \textbf{5.97} & 5.95 & $-0.02$ & Energy-ranked \\
\midrule
Qwen2.5-Coder-0.5B  & 4.07           & \textbf{4.25} & $+0.18$ & smaller baseline \\
\bottomrule
\end{tabular}
\end{table}

\paragraph{A closer look at Qwen2.5-Coder-14B} We examine one model in detail, Qwen2.5-Coder-14B, because it stands as a state-of-the-art open code model across standard benchmarks, making it the ideal candidate for rigorous testing. Crucially, it is also the one mid-size model where runtime ranking reaches a higher aggregate CARET than energy ranking at the fine-tuning stage. This presents the most demanding and conservative setting for an energy-first procedure, ensuring our approach is tested under the strictest possible conditions. Table~\ref{tab:main-results} reports its configurations in detail.

\begin{table*}[t]
\centering
\small
\caption{Baseline results for Qwen2.5-Coder-14B. Bold marks best per column. Energy-SFT ($^{\star}$) is advanced to RL.}
\label{tab:main-results}
\begin{tabular}{lrrrrrrr}
\toprule
Configuration & Compile\% & Tests\% & Valid\% & Mean ERR\% & Med.\ ERR\% & Beat-GT\% & CARET\% \\
\midrule
Qwen14B-Base-ZS   & 1.5  & 0.5  & 0.6   & 0.47   & 0.66    & 0.0   & 0.00 \\
Qwen14B-Base-GP       & 0.1  & 0.0  & 0.0   & 0.00   & 0.00    & 0.0   & 0.00 \\
Qwen14B-Instr-ZS      & \textbf{78.9} & \textbf{55.2} & \textbf{67.9}  & $-0.43$  & 0.62  & 11.8  & $-2.92$ \\
Qwen14B-Instr-GP      & 72.0 & 43.6 & 58.2  & $-6.03$  & 1.04  & 12.0  & $-4.07$ \\
Qwen14B-Energy-SFT$^{\star}$    & 60.3 & 42.2 & 53.0  & 7.95   & $-0.03$ & 12.7  & 4.45 \\
Qwen14B-Runtime-SFT (PIE)   & 48.0 & 29.8 & 39.2  & \textbf{30.17}  & \textbf{27.42} & \textbf{25.1}  & \textbf{7.64} \\
Qwen14B-SFT-Instr     & 38.0 & 23.7 & 30.8  & 6.39   & 0.28  & 10.1  & 0.19 \\
\bottomrule
\end{tabular}
\end{table*}

\paragraph{Energy-SFT and Runtime-SFT on Qwen-14B} On this model Runtime-SFT reaches a higher standalone CARET than Energy-SFT ($7.64\%$ versus $4.45\%$) and beats the human reference more often ($25.1\%$ versus $12.7\%$). RQ1 anticipates this, since energy and runtime agree closely across problems, so a runtime ranking still selects mostly energy-reducing transformations. The two do not differ in whether they optimize but in how their reductions spread. Energy-SFT behaves as a precision optimizer, reducing energy on fewer outputs but cutting deeper, with a high-reduction mode covering $31.0\%$ of valid outputs at a median $77.2\%$, against Runtime-SFT's more frequent but shallower mode ($63.3\%$ at a median $37.5\%$), and the two address largely the same opportunities. On these short compute-bound problems the largest reductions come through cycle savings, so the two rankings converge on how much energy they save. They diverge on the properties deployment depends on. Energy-SFT compiles $60.3\%$ of its outputs against Runtime-SFT's $48.0\%$, produces fully correct outputs more often ($42.2\%$ versus $29.8\%$), and halves the rate of large energy regression, so $52.9\%$ of its outputs both compile and avoid a large regression against $38.9\%$ for the runtime ranking. Where an optimization must not break working code, energy ranking is the safer choice even though runtime ranking scores higher in aggregate CARET on this model. Energy ranking is also the durable signal RQ1 identifies, since the within-problem power differences a runtime ranking ignores grow on longer workloads. The two rankings thus optimize the same opportunities from different starting points, establishing Energy-SFT as the superior foundation for safe, production-grade optimization.

Fine-tuning establishes energy optimization as a learnable capability, with correctness rather than optimization quality as the constraint that remains. Energy-SFT makes $53.0\%$ of its outputs valid, and the CARET decomposition (Section~\ref{sec:caerr-detail}) shows that, on the outputs that already pass, raising the compile rate returns more than pushing energy reduction further. The reduction it achieves is concentrated rather than uniform, since most competitive-programming solutions already run close to their lowest achievable energy, so the gains come from a high-reduction mode while the typical output changes little. This concentration is a property of the problem set rather than a training weakness, and the aggregate reduction sits well above chance (one-sample Wilcoxon $W = 70{,}484$, Holm-Bonferroni adjusted $p = 6.16 \times 10^{-12}$ across the inferential tests we report, rank-biserial $r = 0.325$). Energy-SFT therefore delivers a statistically supported energy reduction with a safer correctness profile than the runtime baseline, making it the definitive configuration for safe, reliable energy optimization.

\begin{findingbox}[RQ2 answer]
Fine-tuning on energy-contrastive pairs successfully teaches the optimization pattern across models, establishing energy efficiency as a learnable capability. While Energy-SFT behaves as a precision optimizer by cutting energy deeply on fewer outputs, its primary advantage lies in deployment reliability, offering a consistently stronger correctness and safety profile than conventional runtime-ranked baselines. However, correctness remains the binding constraint. With under $55\%$ of outputs compiling to valid programs, supervised fine-tuning alone is insufficient for standalone deployment. Nevertheless, the robust cross-model transferability of this effect establishes Energy-SFT as the definitive foundation for energy-efficient code generation.
\end{findingbox}
\subsection{Answering RQ3: Does Closing the Simulation-in-the-Loop Cycle Improve on Fine-Tuning?}
\label{sec:rq3}

This section answers whether closing a simulation-in-the-loop feedback cycle with reinforcement learning improves on supervised fine-tuning, and whether the improvement holds across the reward signal and the initialization. Closing the loop means generating candidates with the current policy, scoring each one through the compile-test-simulate harness, and updating the policy on the resulting reward (Section~\ref{sec:grpo}). Token-level supervision gives no gradient for whether a program compiles or passes its tests, so the loop's correctness-weighted reward targets the bottleneck that fine-tuning cannot reach.

\paragraph{Exploring initializations and reward configurations.} Closing the loop leaves two design choices open: the fine-tuned checkpoint the policy starts from and the simulation-derived reward it optimizes. The initialization sets the baseline behavior, while the reward provides the signal that reshapes it online. Rather than conducting an exhaustive grid search, we structure our evaluation progressively. We begin by comparing our pure energy approach (Energy-SFT with an energy reward) directly against the strongest prior-art baseline (Runtime-SFT with a runtime reward). Next, because RQ2 showed runtime to be an effective proxy for energy on this model, we explore a cross-objective configuration (Energy-SFT with a runtime reward). Finally, we evaluate the energy-delay product (EDP) reward, which jointly scores energy and runtime in a single signal, connecting back to the superior multi-objective ranking observed in RQ2. We evaluate every checkpoint on the same held-out problems as RQ2, under identical hyperparameters and compute budgets, with paired per-problem comparisons against the supervised starting point. Table~\ref{tab:ablation-grpo} reports the held-out outcome for each configuration, grouped progressively.

\begin{table}[t]
\centering
\small
\caption{Comparison of closed-loop configurations across initialization and reward choices. \textit{Init} denotes the SFT checkpoint. Bold marks the primary configuration used in subsequent RQs.}
\label{tab:ablation-grpo}
\begin{tabular}{lrrrrrrr}
\toprule
Init + Reward & Train~Succ\% & Compile\% & Tests\% & ERR$_{\text{valid}}$\% & Beat-GT\% & SFT~CARET\% & CARET\% \\
\midrule
\multicolumn{8}{l}{\textit{Pure single-axis vs.\ baseline}} \\
E-SFT + energy (Ours) & \textbf{53.4} & 82.0 & 51.1 & \textbf{30.71} & \textbf{33.4} & 4.45 & \textbf{12.78} \\
Runtime-SFT + runtime (Afterburner) & 49.3 & \textbf{88.5} & \textbf{64.6} & 23.37 & 22.1 & \textbf{7.64} & 12.19 \\
\midrule
\multicolumn{8}{l}{\textit{Cross-objective exploration}} \\
E-SFT + runtime & 53.2 & 82.2 & 51.1 & 30.54 & 33.3 & 4.45 & 14.44 \\
\midrule
\multicolumn{8}{l}{\textit{Multi-objective (Energy-Delay Product)}} \\
\textbf{E-SFT + EDP} & \textbf{63.5} & 81.7 & 44.6 & 34.20 & \textbf{58.4} & 4.45 & 12.63 \\
Runtime-SFT + EDP & 34.6 & 86.3 & 40.8 & \textbf{46.93} & 51.5 & \textbf{7.64} & \textbf{16.41} \\
\bottomrule
\end{tabular}
\end{table}

\paragraph{Every closed-loop configuration decisively beats fine-tuning.} All five configurations land in a $4.22\,$pp CARET band ($12.19$ to $16.41\%$), and even the weakest exceeds the strongest fine-tuning-only variant on this model, the energy-delay supervised variant at $11.02\%$, clearing the runtime-ranked baseline at $7.64\%$ by $4.55\,$pp.

\paragraph{Our approach overcomes a weaker starting point to beat the baseline.} RQ2 leaves the runtime fine-tune well ahead of the energy fine-tune at the supervised stage ($7.64\%$ against $4.45\%$ CARET). By deliberately choosing this weaker performing energy checkpoint as our initialization, we stress-test our approach against a difficult starting point. The closed loop begins with our energy initialization $3.19\,$pp behind on CARET. Giving each initialization its aligned reward—the energy fine-tune an energy reward and the runtime fine-tune a runtime reward—overturns that order. The runtime-aligned model reproduces the test-time runtime-focused GRPO methodology of Afterburner~\cite{du2025afterburner} inside our protocol. Despite intentionally starting from this poorer performing supervised checkpoint, our energy-aligned model comes out ahead of this baseline on both primary metrics, reaching $12.78\%$ aggregate CARET against the baseline's $12.19\%$, and a wide margin on the rate of beating the human reference ($33.4\%$ against $22.1\%$). 

This reversal prompted us to explore a cross-objective setup: E-SFT combined with a runtime reward. Because energy-ranked training already learns the cycle-driven transformations that a runtime objective rewards (RQ2), combining the robust energy starting point with a runtime reward yields a high CARET of $14.44\%$. This validates the flexibility of the E-SFT checkpoint, proving it is highly adaptable when optimizing for execution time as an additional proxy route to lower energy.

\paragraph{The energy-delay reward carries the human-beating rate.} Pairing either initialization with an energy-delay reward changes the picture on the human-beating axis. The two energy-delay cells reach the highest rates of beating the human reference ($58.4\%$ from the energy initialization and $51.5\%$ from the runtime one), roughly double the $33.4\%$ and $22.1\%$ of their single-axis counterparts, and the runtime-initialized cell also lifts aggregate CARET to the top of the band ($16.41\%$, from $12.19\%$ with a runtime reward). The energy-delay reward therefore carries the human-beating rate, while it leaves aggregate CARET near the rest of the band on the energy initialization ($12.63\%$ against $12.78\%$). This mirrors RQ2, where combining time and energy into a single sort produced the strongest supervised variant, with the same combination now acting through the reward rather than a static ranking.

The CARET margins separating the five cells are small relative to single-seed run-to-run variation, so we treat the band as one robust result and read the within-band comparisons as directional rather than exact. The rate of beating the human reference is the exception, diverging by $36.3\,$pp across the five cells, an order of magnitude wider than the CARET spread, and it tracks the reward identity rather than the initialization. The simulation-in-the-loop cycle is therefore the dominant mechanism for aggregate CARET, while the energy-delay reward is what carries the human-beating rate.

\begin{figure}[t]
\centering
\includegraphics[width=\columnwidth]{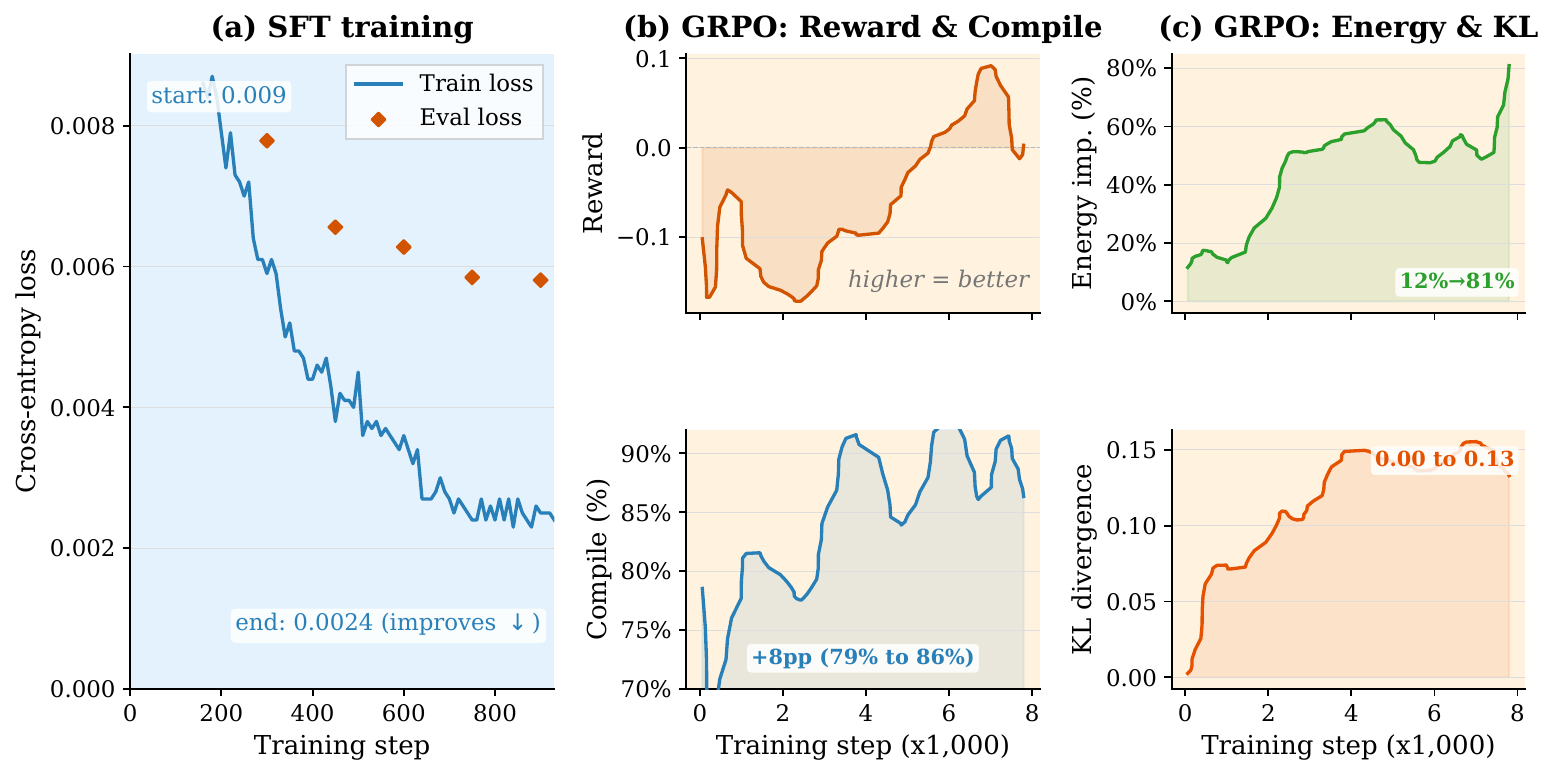}
\caption{SFT (left) and GRPO (right) training progression. Panel (a) illustrates SFT loss reduction. Panels (b) and (c) demonstrate the closed-loop effectiveness: the model rapidly learns to emit correct code (compile rate rises) and optimize energy (energy-improvement rate climbs from 12\% to over 80\%), confirming that the reward signal successfully teaches deep energy optimization.}
\label{fig:training-dynamics}
\end{figure}

\paragraph{The energy-first configuration in detail.} For the detailed view we follow the energy-initialized, energy-delay-rewarded configuration (E-SFT $+$ EDP). This configuration integrates the most holistic simulation feedback, serving as the definitive energy-first design. Figure~\ref{fig:training-dynamics} traces the training progression. As shown in Figure~\ref{fig:training-dynamics}(b) and (c), the simulation feedback actively reshapes model behavior. The reward ranges over $[-1.0, 1.5)$, so the mean reward improving toward zero reflects failing rollouts giving way to correct ones. Crucially, Figure~\ref{fig:training-dynamics}(c) shows the energy-improvement rate rising steeply from 12\% to 81\%, definitively demonstrating that the model learns to isolate and optimize energy patterns throughout the reinforcement phase. Simultaneously, the training compile rate gains about 8\,pp (Figure~\ref{fig:training-dynamics}(b)), while a transient rise in KL divergence near the end recovers quickly without negatively affecting the final checkpoint.

\paragraph{Closing the loop improves every primary metric.} The compilation rate rises from 60.3\% to 81.7\%, an increase of 21.4\,pp that sharply reduces the fraction of wasted outputs, and CARET rises from 4.45\% to 12.63\%, a 2.84$\times$ improvement that clears the 7.64\% runtime-ranked fine-tuning ceiling on this model (Section~\ref{sec:rq2}). Optimization quality rises with it. Mean ERR on valid outputs (ERR$_{\text{valid}}$) rises from 7.95\% to 34.20\%, a 4.3$\times$ gain, while Beat-GT, the fraction of valid outputs that match or exceed the per-problem human-expert reference, rises from 12.7\% to 58.4\% and the overall valid rate from 53.0\% to 63.5\%. A paired Wilcoxon signed-rank test on the 97 problems where both stages produce valid outputs confirms the per-problem gain is not the work of a few outliers ($W{=}482$, $p{=}9.3 \times 10^{-12}$, rank-biserial $r{=}0.90$, a large effect), and under Holm-Bonferroni correction across the nine inferential tests the adjusted $p{=}6.5 \times 10^{-11}$, so the gain survives multiple-comparison correction. A bootstrap 95\% CI on per-sample ERR improvement spans $[7.89\%, 291.59\%]$ on the paired set. The wide upper bound reflects a small number of problems on which the closed loop achieves order-of-magnitude algorithmic gains (cycle reductions of 30$\times$ or more) while fine-tuning achieves near-zero ERR, so the interval is a property of the heavy-tailed improvement distribution rather than a unit error. The decomposition in Section~\ref{sec:caerr-detail} attributes the dominant share of the gain to this rise in ERR$_{\text{valid}}$, with the correctness improvements carrying it to a larger share of outputs.

\paragraph{The correctness picture is more nuanced than the headline rates.} The overall test pass rate improves modestly, from 42.2\% to 44.6\%, while the conditional pass rate, the fraction of compiled outputs that also pass all tests, decreases from 70.0\% to 54.6\%. The conditional rate is misleading in isolation, because the compiled-output denominator grows by 36\% under the closed loop, so in absolute terms it produces more test-passing outputs, roughly 545 against 514 for fine-tuning. The rate falls because the closed loop compiles programs fine-tuning could not, then attempts harder transformations on them. This shift toward deeper transformations that are harder to get fully correct is consistent with the Shypula et al.~\cite{shypula2023pie} finding that conditioning generation on the maximum performance target alone constrains the space of reachable correct solutions, while mixing lower score targets with the maximum improves correctness, and the closed loop's reward structure applies an analogous maximum-optimization pressure. CARET is the appropriate aggregate comparison, since it folds every failure into one deployment-relevant metric, and its 2.84$\times$ gain combines a 4.6\% rise in absolute test-passing count with a 4.3$\times$ rise in ERR$_{\text{valid}}$, so the bulk of the improvement comes from optimization quality on valid outputs rather than from correctness alone. Section~\ref{sec:rq4} classifies these transformations and quantifies the shift directly.

\begin{figure}[t]
\centering
\includegraphics[width=0.65\columnwidth]{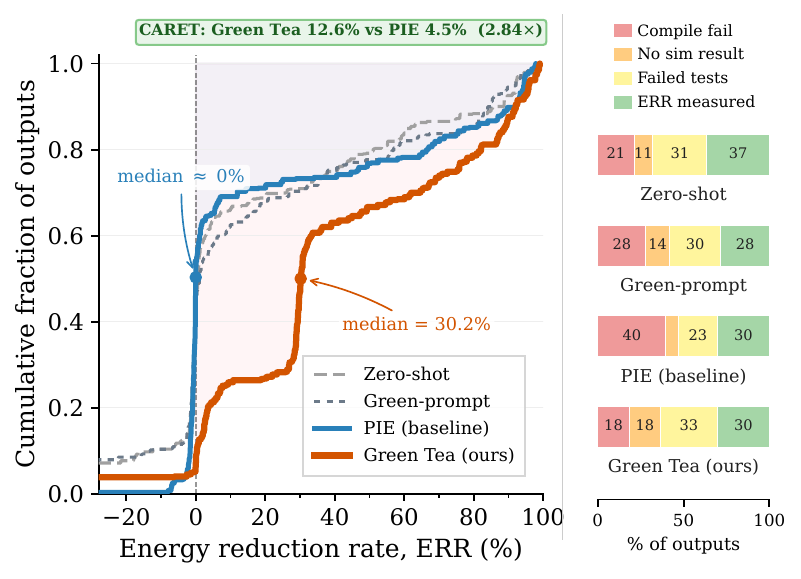}
\caption{Empirical Cumulative Distribution Function (CDF) of ERR across training stages (valid outputs only). The right panel categorizes outputs. GRPO stochastically dominates SFT, collapsing the zero-gain shelf.}
\label{fig:hero-progression}
\end{figure}

\begin{figure}[t]
\centering
\includegraphics[width=0.90\columnwidth]{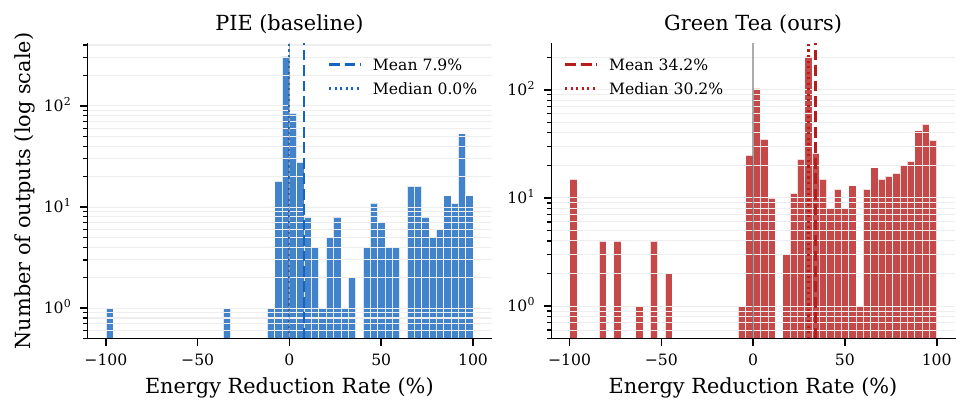}
\caption{ERR distribution for Energy-SFT vs.\ Green Tea (Energy-SFT+GRPO). Green Tea collapses the zero-gain shelf and shifts the distribution toward positive energy savings.}
\label{fig:err-dist}
\end{figure}

\paragraph{The loop shifts the whole reduction distribution.} The ERR distribution shifts qualitatively (Figure~\ref{fig:hero-progression}). The fraction of valid outputs with negative ERR falls from 51.5\% to 7.2\%, and the interquartile range moves from $[-0.7\%, 46.4\%]$ to $[7.7\%, 75.1\%]$. As shown in Figure~\ref{fig:err-dist}, the bimodal shelf near zero for Energy-SFT collapses into a concentrated positive distribution under Green Tea. At the per-problem level, 74.1\% of test problems achieve positive CARET under the closed loop versus 40.6\% under fine-tuning, with median per-problem CARET rising from 0.00\% to 7.04\%, so the closed loop shifts the entire distribution rightward, not just the mean.

\subsubsection{Decomposing the CARET Gain}
\label{sec:caerr-detail}
\begin{figure}[t]
\centering
\includegraphics[width=0.70\columnwidth]{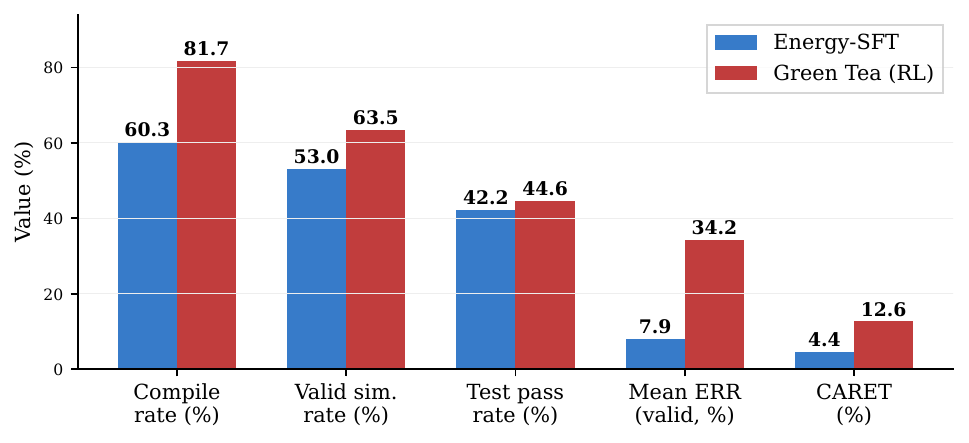}
\caption{CARET decomposition across Energy-SFT and Green Tea (Energy-SFT+GRPO). CARET is factored into compilation, simulation, test pass rates, and mean ERR.}
\label{fig:caerr}
\end{figure}

To precisely locate the source of GRPO's performance gains, we factor CARET into its constituent components (Figure~\ref{fig:caerr}). We introduce the valid simulation rate alongside compile and test pass rates to track how the correctness gate filters executable programs. This factorization is an approximation because its factors correlate. It underestimates CARET by $1.30\times$ for SFT (where correct outputs are typically more efficient) and overestimates it by $1.22\times$ for GRPO (where the correctness gate actively filters out high-variance, extreme-reduction transforms). We therefore treat the factorization interpretively, pairing individual factors alongside aggregate CARET. A counterfactual analysis isolates the primary driver of the gain. Holding SFT's ERR$_{\text{valid}}$ fixed at $7.95\%$ while substituting GRPO's test pass rate ($44.6\%$) yields a counterfactual CARET of $0.446 \times 7.95\% = 3.54\%$, strictly below SFT's actual $4.45\%$. Thus, isolated correctness improvements would have actually lowered CARET. Instead, GRPO's massive $2.84\times$ aggregate gain derives entirely from the $4.3\times$ surge in ERR$_{\text{valid}}$ (from 7.95\% to 34.20\%), which the correctness improvements subsequently propagate across a broader set of outputs. Furthermore, this gain is broad-based rather than outlier-driven. Per-problem test pass rates correlate only weakly with energy reduction rates ($r{=}0.024$ for SFT, $r{=}0.131$ for GRPO), proving that solving a problem correctly and solving it efficiently remain largely independent challenges.

\paragraph{What the initialization and the reward each contribute.} The progressive evaluation separates the two levers: the supervised initialization and the on-policy reward. The two energy-delay (EDP) cells make this split concrete. The runtime-initialized one carries the higher aggregate CARET and the higher mean ERR on valid outputs ($46.93\%$), while the energy-first one carries the higher Beat-GT and the higher fractional test pass rate ($44.6\%$ against $40.8\%$), so the choice between them is a deployment-context choice rather than a strict quality ranking. Two mechanisms explain why the rewards converge on aggregate CARET yet diverge on the human-beating rate. First, KL regularization ($\beta = 0.04$) bounds how far the policy drifts from its supervised reference, so when the on-policy reward signal is weak the inference behavior stays near the supervised checkpoint and the supervised-stage differences dominate the held-out metric. Second, while many successful optimizations naturally improve both execution time and energy, specifically including energy feedback in the reward keeps the model firmly aligned with energy outcomes. The wider Beat-GT divergence highlights the strength of the energy-delay reward: its gradient amplification strongly reinforces cycle-aligned wins—where a 2$\times$ improvement in both energy and cycles yields a 4$\times$ ratio—allowing it to surface high-quality, reference-beating transformations far more reliably than runtime training alone.

\paragraph{The headline reduction is robust.} The 12.63\% figure does not depend on how measurement failures are scored or how outputs are aggregated. Under the strictest convention, where only fully-correct outputs contribute non-zero ERR, CARET falls to 5.62\%. Measurement failures are rare under the closed loop, so the most pessimistic convention, which counts compiled-but-unmeasured outputs as $\text{ERR}{=}-0.5$, leaves CARET at 12.63\%, unchanged from the default, and the headline therefore lies between 5.62\% and 12.63\% across the conventions we examine. It is also stable to the aggregation unit, reaching 12.26\% when each base solution is weighted once and 13.78\% per problem, so it does not rest on the per-pair denominator. The 5.62\% figure is the one comparable to binary-gated efficiency benchmarks, which credit only fully-correct outputs, and we report the graded CARET as primary because it also credits the partial progress of near-correct high-reduction outputs, pairing it throughout with the convention-independent compile and human-beating rates.

\paragraph{The gain tracks the training signal, not scale.} A per-pair comparison places the closed-loop 14B model against larger off-the-shelf references. Under a per-problem evaluation, Qwen2.5-Coder-32B and DeepSeek-Coder-V2-Lite zero-shot appear to reach a higher aggregate CARET, about 15\%, but that denominator credits each model with a single favorable baseline per problem. Because the deployment task is to improve a given program rather than to select the best of several candidates, the fair comparison is the per-pair denominator that scores every base-target pair, matching the fine-tuned and reinforcement-learning rows. On it, Qwen2.5-Coder-32B zero-shot reaches $-16.19\%$ CARET while compiling 93\% of its outputs, a genuine regression toward more energy-costly code, and DeepSeek-Coder-V2-Lite zero-shot reaches 4.27\%. Thus, the closed-loop 14B model's 12.63\% exceeds both larger references by roughly 8 to 29\,pp on the same footing. The advantage therefore comes directly from the targeted training signal rather than from raw scale. To rigorously validate our findings and ensure the simulation-in-the-loop methodology generalizes beyond a single model family, we replicated the complete pipeline on an entirely different architecture: DeepSeek-Coder-6.7B. Despite a weak supervised starting point on this model (Section~\ref{sec:rq2}), the closed loop successfully lifted its CARET to $5.99\%$. This independent replication confirms that our approach drives energy optimization across different model architectures, significantly strengthening the generalizability of the core claim.

\paragraph{Transfer across model size.} The closed loop's contribution is not uniform across model sizes. Applying the same two-stage procedure to Qwen2.5-Coder-7B, initializing reinforcement learning from the 7B energy fine-tune deepens the optimization on the outputs that compile, raising mean energy reduction on valid outputs from 35.03\% to 38.26\% and the rate of beating the human reference from 28.1\% to 30.5\%. The compile rate stays at its fine-tuning level (64.0\% to 64.6\%), and over the problems both stages cover the closed loop and fine-tuning do not differ on aggregate CARET (paired Wilcoxon $p{=}0.12$). This contrasts with the 14B model, where the loop lifts the compile rate from 60.3\% to 81.7\% and CARET from 4.45\% to 12.63\%. The optimization-depth benefit of the closed loop therefore reliably transfers across model sizes, while the correctness-recovery benefit that drives the massive 14B aggregate gain is tied to model capacity rather than being a property the closed loop guarantees at every scale.

\paragraph{Behavior across sampling budgets.} The headline metrics use a single low-temperature generation, so a final check asks whether the closed loop's advantage survives a larger inference budget. To prevent any loophole regarding temperature settings, we evaluated stochastic sampling ($n{=}10$, $T{=}0.8$) across all 143 test problems. Under a best-of-16 budget at $T{=}0.8$, scoring each problem by the best test-pass-weighted reduction across its 16 samples and counting non-compiling or failing samples as zero, the closed loop holds its CARET lead, reaching $33.63\%$ against $24.36\%$ for fine-tuning. That $9.3\,$pp gap tracks the $8.2\,$pp gap at a single sample ($12.63\%$ against $4.45\%$), confirming that a larger sampling budget does not let fine-tuning close the energy advantage. While pass@10 regresses modestly (from 48.95\% to 46.15\%), pass@1 improves from 28.18\% to 31.40\% (Table~\ref{tab:passk}). The pass@10 drop occurs because the aggressive refactorings the closed loop favors occupy a narrower space than fine-tuning's outputs, surfacing fewer orthogonal correct solutions. However, sampling and the reinforcement stage act as complementary rather than substitutable gains, making the closed loop the stronger choice across inference budgets, reliably delivering energy reductions that fine-tuning cannot reach alone.

\begin{table}[t]
\centering
\caption{Pass@k correctness at temperature 0.8, $n{=}10$. Conditional ERR is unavailable because energy simulations were not run on stochastic samples.}
\label{tab:passk}
\begin{tabular}{lrrr}
\toprule
Method & pass@1 & pass@5 & pass@10 \\
\midrule
SFT & 28.18 & 42.29 & 48.95 \\
SFT$+$GRPO & 31.40 & 41.47 & 46.15 \\
\bottomrule
\end{tabular}
\end{table}

\begin{findingbox}[RQ3 answer]
Closing the simulation-in-the-loop cycle improves on fine-tuning, raising CARET from 4.45\% to 12.63\% (a 2.84$\times$ gain) and the rate of beating the human reference from 12.7\% to 58.4\% on the energy-first configuration. The closed loop itself drives the massive aggregate CARET gain, reliably elevating performance across all evaluated configurations. The reward identity matters decisively for beating the human reference, where the energy-delay configurations roughly double the baseline rate. Thus, the energy-delay reward fully earns its added McPAT simulation cost by surfacing expert-quality, cycle-aligned energy optimizations that a simpler reward misses.
\end{findingbox}
\subsection{Answering RQ4: Which Optimization Strategies Does the Training Induce?}
\label{sec:rq4}

The preceding sections quantify \emph{how much} energy the model saves; this section examines \emph{how} it saves it. We analyze the generated programs directly to answer three questions: which source-level transformations the training induces, whether those transformations are the genuine source of the savings, and whether the model successfully navigates the IPC trap that causes runtime proxies to fail. To do this, we classify each transformation into a three-tier taxonomy and track the distribution shift across training stages. We then explicitly rule out superficial fast input-output substitution and compiler optimization as the primary drivers of the gains, verify that the improvements reach the difficult IPC-trap regime, and probe the attention mechanism to understand which input cues the model relies on.

\paragraph{A transformation taxonomy.}
To understand how the model achieves its gains, we require a taxonomy of code transformations. While our recent work~\cite{mehditabar2026validated} identifies 12 granular categories of root causes across a validated taxonomy of software energy smells, evaluating an LLM's true capability requires measuring the \emph{optimization depth} of its generated code transformations. Therefore, we group these granular smells into a three-tier macro-classification (Table~\ref{tab:taxonomy}) rooted in fundamental performance-engineering principles~\cite{bentley1982writing,leiserson2020there}: asymptotic-complexity changes (Category~A), constant-factor changes at the same complexity class (Category~B), and cosmetic edits with no execution effect (Category~C). This three-tier taxonomy is strictly empirical rather than stylistic; the categories reliably order by measured retired-instruction count reductions (Section~\ref{sec:rq4-compiler}), providing an objective metric of optimization depth. To scale this analysis across thousands of generated outputs, we assign each valid program to a category using a deterministic keyword heuristic. While less sophisticated than a full semantic AST parser, this heuristic's validity is empirically confirmed by the strict monotonic relationship between its assigned categories and actual retired-instruction reductions. Specifically, if an output fundamentally alters looping structures or swaps standard library data structures (e.g., \texttt{vector} to \texttt{set}), it is flagged as Category~A. If it introduces fast I/O (\texttt{scanf}), preallocation (\texttt{reserve}), or converts heap allocations to stack arrays without changing the algorithmic core, it is classified as Category~B. All other valid edits, such as type aliases or formatting changes, default to Category~C. We release these per-output assignments for independent audit.

\begin{table}[ht]
\centering
\caption{Optimization taxonomy for model-generated code transformations.}
\label{tab:taxonomy}
\begin{tabular}{clp{7.5cm}}
\toprule
Category & Label & Description \\
\midrule
A & Complexity-class change & Fundamental algorithm swap (for example, $O(n^2) \to O(n \log n)$), data structure replacement (for example, \texttt{set} $\to$ \texttt{unordered\_set}), or mathematical reformulation that reduces asymptotic complexity. \\
B & Constant-factor optimization & Loop restructuring at the same complexity class, fast I/O substitution (\texttt{scanf}/\texttt{printf} over \texttt{cin}/\texttt{cout}), memory layout changes (array versus vector), compiler pragmas, or microarchitectural tuning. \\
C & Cosmetic or unclassified & Variable renaming, formatting changes, type aliases, comment additions, structurally identical code, or any edit not matched to Category~A or~B. \\
\bottomrule
\end{tabular}
\end{table}

Four distinct patterns dominate the fine-tuned model's valid outputs. First, cosmetic type aliases (e.g., \texttt{typedef long long ll}) are common Category~C conventions that yield no energy benefit. Second, fast input-output substitution (e.g., \texttt{scanf}/\texttt{printf}) and third, memory-layout changes (e.g., replacing a heap-allocated \texttt{vector} with a stack array) are the primary Category~B transforms that effectively cut overhead. Finally, fundamental algorithmic shifts are the Category~A cases, representing $3.9\%$ of valid outputs. To see how these play out in practice, we examine two worked examples: one showcasing algorithmic simplification, and another illustrating the IPC trap through memory-layout optimization.

\begin{figure}[t]
\centering
\begin{minipage}[t]{0.33\linewidth}\centering{\footnotesize\textbf{Baseline}}\par\vspace{2pt}
\begin{lstlisting}[style=cppbox, basicstyle=\ttfamily\scriptsize, numbers=none, breaklines=true]
#include <bits/stdc++.h>
using namespace std;
long long int ar[1000000];
int main() {
    long long int x, y, rem, temp, i, j, total, f = 0;
    cin >> x >> y;
    temp = x;
    for(i = 0; i < 1000000; i++) ar[i] = -1;
    i = 0;
    while(temp >= 1) {
        rem = temp % y; temp /= y;
        ar[i] = rem; i++;
    }
    for(j = 0; j < i; j++) { total = ar[j] % 10; f++; }
    cout << f << endl;
}
\end{lstlisting}
\end{minipage}\hfill
\begin{minipage}[t]{0.33\linewidth}\centering{\footnotesize\textbf{SFT (ERR 96.7\%)}}\par\vspace{2pt}
\begin{lstlisting}[style=cppbox, basicstyle=\ttfamily\scriptsize, numbers=none, breaklines=true]
#include <iostream>
using namespace std;
int main() {
    long long int x, y;
    cin >> x >> y;
    long long int count = 0;
    while(x > 0) { 
        count++; x /= y; 
    }
    cout << count << endl;
    return 0;
}
\end{lstlisting}
\end{minipage}\hfill
\begin{minipage}[t]{0.33\linewidth}\centering{\footnotesize\textbf{GRPO (Rejected)}}\par\vspace{2pt}
\begin{lstlisting}[style=cppbox, basicstyle=\ttfamily\scriptsize, numbers=none, breaklines=true]
#include <iostream>
using namespace std;
int main() {
    long long int x, y;
    cin >> x >> y;
    long long int sum = 0;
    while(x > 0) { 
        sum += x % y; x /= y; 
    }
    cout << sum << endl;
    return 0;
}
\end{lstlisting}
\end{minipage}
\caption{\textbf{Example~1 (p02766, Category~A, algorithmic simplification).} SFT simplifies the algorithm, removing array allocations for a 96.7\% ERR. The GRPO output achieves an even higher nominal 97.7\% ERR by summing digits instead of counting iterations, but is successfully suppressed by the correctness gate for failing tests.}
\label{fig:ex-p02766}
\end{figure}

The first example (Figure~\ref{fig:ex-p02766}) counts how many digits $x$ has in base $y$. The baseline inefficiently allocates and initializes an 8\,MB global array to hold the digits before counting them. The fine-tuned (SFT) output simplifies the algorithm to count loop iterations directly, entirely removing the array and driving a $96.7\%$ energy reduction. Notably, on this same problem, the closed-loop GRPO policy produced a compile-valid output that incorrectly summed the digits rather than counting them. Although this incorrect output achieved an even higher $97.7\%$ nominal ERR, it failed 102 of 104 test cases and was thus zeroed out by the correctness gate. This perfectly illustrates the reward gate in action, successfully suppressing a broken high-ERR transformation.

The closed loop also performs deep structural refactorings. On a bitmask-reachability problem (p03395), it produced a correct non-trivial optimization, replacing a precomputed power table and its initialization loop with direct bit-shifts while leaving the core reachability search over the candidate set intact; the result passed all $106$ test cases, reduced energy by $28.3\%$, and beat the human-optimized reference by $8.4$ percentage points. It even ventures into advanced data structures: on a range-query problem (p02599) it generated a sophisticated Fenwick-tree solution with offline query processing that reached $86.8\%$ energy reduction.

\begin{figure}[t]
\centering
\begin{minipage}[t]{0.33\linewidth}\centering{\footnotesize\textbf{Baseline}}\par\vspace{2pt}
\begin{lstlisting}[style=cppbox, basicstyle=\ttfamily\scriptsize, numbers=none, breaklines=true]
#include <iostream>
using namespace std;
int main() {
    int n, a[10005]={}, cnt=0;
    cin >> n;
    for(int i=0; i<n; i++) {
        cin >> a[i];
        while(a[i]%2 == 0) { 
            cnt++; a[i] >>= 1; 
        }
    }
    cout << cnt << endl;
}
\end{lstlisting}
\end{minipage}\hfill
\begin{minipage}[t]{0.33\linewidth}\centering{\footnotesize\textbf{SFT (ERR 11.9\%)}}\par\vspace{2pt}
\begin{lstlisting}[style=cppbox, basicstyle=\ttfamily\scriptsize, numbers=none, breaklines=true]
#include <iostream>
using namespace std;
int main() {
    int n, ans = 0;
    cin >> n;
    for(int i=0; i<n; i++) {
        int a; cin >> a;
        while(a%2 == 0) { 
            a /= 2; ans++; 
        }
    }
    cout << ans << endl;
}
\end{lstlisting}
\end{minipage}\hfill
\begin{minipage}[t]{0.33\linewidth}\centering{\footnotesize\textbf{GRPO (ERR 39.0\%)}}\par\vspace{2pt}
\begin{lstlisting}[style=cppbox, basicstyle=\ttfamily\scriptsize, numbers=none, breaklines=true]
#include <cstdio>
using namespace std;
int main() {
    int n, ans = 0;
    scanf("%d", &n);
    for(int i=1; i<=n; i++) {
        int x; scanf("%d", &x);
        while(x%2 == 0) {
            ans++; x /= 2;
        }
    }
    printf("%d\n", ans);
}
\end{lstlisting}
\end{minipage}
\caption{\textbf{Example~2 (p03325, Category~B, memory layout \& I/O).} SFT removes the array (IPC trap, 11.9\% ERR). GRPO stacks fast I/O on top of it, reaching 39.0\% ERR and $1.64\times$ speedup.}
\label{fig:ex-p03325}
\end{figure}

The second example (Figure~\ref{fig:ex-p03325}) counts the right-shifts needed to make all array elements odd. The baseline reads every value into a 40\,KB stack array, whereas the fine-tuned output processes each value inline. Because the computation loop dominates execution time, runtime improves only modestly ($1.13\times$). However, energy improves significantly more ($11.9\%$) because eliminating the array directly reduces cache pressure and per-cycle memory traffic. This is a classic instance of the IPC trap, where the energy savings significantly exceed the runtime speedup. The closed loop optimizes this same problem even further, reaching a $39.0\%$ ERR and a $1.64\times$ speedup while keeping all tests passing. It achieves this by adopting \texttt{scanf}/\texttt{printf} fast input-output alongside the inline processing, demonstrating how the closed loop successfully stacks multiple Category~B constant-factor optimizations to maximize energy savings.

\paragraph{Closing the loop shifts cosmetic edits to constant-factor optimizations.}
Table~\ref{tab:category-err} reports ERR by optimization category for the valid fine-tuning and closed-loop outputs. The closed loop dramatically redistributes the model's strategies: cosmetic Category~C edits plummet by $43.9\,$pp, while substantive Category~B constant-factor optimizations surge by $44.2\,$pp. A Stuart--Maxwell test for paired marginal homogeneity confirms this structural shift is highly significant ($T{=}28.03$, $p{=}8.19 \times 10^{-7}$). This redistribution occurs because the correctness gate effectively suppresses cosmetic outputs (which earn near-zero reward), while the simulation reward heavily amplifies constant-factor transformations. Not only does the loop increase the volume of Category~B edits, it also improves their quality, raising the median ERR for this category from $24.5\%$ to $31.0\%$. This behavior holds across all five evaluated configurations (Table~\ref{tab:ablation-categories}), proving that the closed loop itself, rather than any single reward signal, drives this optimization capability. The shift does not, however, extend to Category~A complexity-class changes, which remain roughly constant. Because algorithmic refactorings are inherently riskier and more likely to fail tests (as seen in Figure~\ref{fig:ex-p02766}), the correctness gate suppresses these high-variance attempts. Consequently, the closed loop achieves its massive aggregate gains by systematically broadening and perfecting reliable constant-factor coverage, rather than gambling on brittle complexity-class refactorings that clear the correctness gate too rarely to move the aggregate.

\begin{table}[t]
\centering
\caption{ERR by optimization category for Energy-SFT and Energy-SFT + EDP. Category~A reports the median due to low sample size outliers.}
\label{tab:category-err}
\begin{tabular}{lrrrr}
\toprule
 & \multicolumn{2}{c}{SFT} & \multicolumn{2}{c}{SFT+GRPO} \\
\cmidrule(lr){2-3}\cmidrule(lr){4-5}
Category & $n$ (\%) & Median ERR\% & $n$ (\%) & Median ERR\% \\
\midrule
A: Complexity-class change & 25 (3.9\%) & 58.1 & 28 (3.6\%) & 54.6 \\
B: Constant-factor optimization & 155 (24.0\%) & 24.5 & 529 (68.2\%) & 31.0 \\
C: Cosmetic or unclassified & 466 (72.1\%) & $-0.1$ & 219 (28.2\%) & 5.8 \\
\bottomrule
\end{tabular}
\end{table}

\begin{table}[t]
\centering
\small
\caption{Optimization category distribution across GRPO configurations (valid outputs only). medA and medB denote the median ERR within their respective categories.}
\label{tab:ablation-categories}
\begin{tabular}{lrrrrrr}
\toprule
\textbf{Config} & \textbf{Valid N} & \textbf{Cat A\%} & \textbf{medA\%} & \textbf{Cat B\%} & \textbf{medB\%} & \textbf{Cat C\%} \\
\midrule
E-SFT + EDP        & 776 &  \textbf{3.6} & 54.6 & 68.2 & 31.0 & \textbf{28.2} \\
R-SFT + EDP        & 790 &  1.6 & \textbf{86.1} & \textbf{68.4} & \textbf{59.5} & 30.0 \\
E-SFT + energy     & \textbf{803} &  2.9 & 60.9 & 56.8 & 28.7 & 40.3 \\
R-SFT + runtime    & \textbf{820} &  1.0 & 12.0 & 35.0 & 28.8 & 64.0 \\
E-SFT + runtime    & 790 &  3.4 & 60.8 & 60.0 & 28.5 & 36.6 \\
\bottomrule
\end{tabular}\end{table}

\paragraph{Reaching the IPC-trap regime.}
As established in RQ1, runtime proxies systematically fail to optimize energy due to the IPC trap, a difficult architectural regime where memory stalls decouple energy from execution time. To prove that our generated gains are not merely exploiting easy, highly-correlated non-trap problems, we evaluate the model's performance within this challenging regime. The test set accurately reflects this difficulty, containing a 67.1\% prevalence of IPC-trap problems ($N{=}143$) that closely matches the 67.8\% baseline rate from RQ1. Under supervised fine-tuning alone, performance sharply diverges: the model struggles significantly more with trap-prone problems, achieving lower compilation success (55.7\% versus 73.3\%, $p < 0.001$) and generating far less aggregate energy reduction (2.81\% CARET versus 9.01\% for non-trap). Closing the simulation loop, however, fundamentally bridges this gap by driving massive simultaneous enhancements. For IPC-trap problems, CARET jumps to 8.19\% (a $2.91\times$ improvement), while non-trap CARET rises to 25.07\% (a $2.78\times$ improvement). Corresponding surges in compilation success (reaching 78.1\% and 91.6\%, respectively) push the human-beating optimization rates to 52.1\% for trap problems and 71.6\% for non-trap problems. While the non-trap regime remains inherently easier to optimize (yielding a statistically higher CARET distribution, Mann--Whitney $p{=}0.0047$), the closed loop successfully nearly triples aggregate performance in both categories. This proves that while the IPC trap persists as a fundamental structural barrier, the closed loop empowers the model to successfully discover deep energy-saving transformations across completely disparate architectural boundaries.

\paragraph{Gains persist beyond fast input-output substitution.}
Fast input-output substitution (\texttt{scanf}/\texttt{printf}) is prevalent, appearing in 59.9\% of valid closed-loop outputs. However, this surface transformation is auxiliary; 91.0\% of these cases also achieve cycle reductions exceeding 15\%, indicating that it typically accompanies deeper algorithmic or structural optimizations. Excluding these outputs yields a CARET of 5.51\%, which remains substantially higher than the fine-tuning baseline of 4.45\%, demonstrating that the gains persist independent of I/O specific changes.

\paragraph{The role of compiler optimization.}
\label{sec:rq4-compiler}

To ensure our measured gains are genuine, all evaluations throughout this work use \texttt{g++ -O3}. We impose this maximum optimization level to set a rigorously high bar: modern compilers automatically perform loop unrolling, constant propagation, and vectorization. By evaluating at \texttt{-O3}, we guarantee that the model is forced to discover deep algorithmic and structural improvements that escape the compiler's reach, rather than merely suggesting trivial refactorings that the compiler would have optimized anyway.

If the compiler were simply masking the differences between baseline and generated code, energy reductions would collapse to zero. Instead, we observe a substantial 12.63\% mean CARET. These energy reductions strongly correlate with strict decreases in dynamic instruction count (Spearman $\rho = -0.947$), with Category~A algorithmic and Category~B constant-factor refactorings yielding 64.0\% and 21.5\% fewer retired instructions, respectively. Furthermore, an optimization-level sweep across 30 stratified problems confirms that the model's benefits are remarkably robust: mean ERR remains completely stable across the compiler stack, yielding 34.23\% at \texttt{-O0} versus 34.38\% at \texttt{-O3} (Pearson $r = 0.995$). The absolute persistence of these gains at maximum optimization levels proves conclusively that the closed loop's source-level transformations provide fundamental performance benefits that far exceed the capabilities of the compiler alone.

\begin{figure}[t]
\centering
\includegraphics[width=0.70\columnwidth]{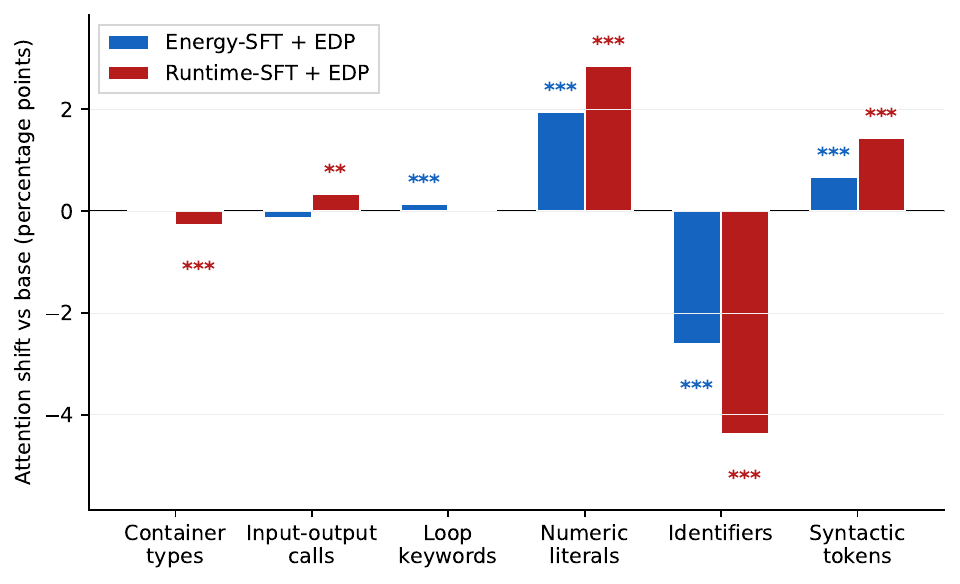}
\caption{Per-input-class attention shift relative to the base model for Energy-SFT and Runtime-SFT (both EDP-rewarded). Stars indicate Wilcoxon significance ($*\,p<0.05$, $**\,p<0.01$, $***\,p<0.001$).}
\label{fig:attn-class-shift}
\end{figure}

\paragraph{What the model attends to.}\label{sec:rq4-attention}
The categorical redistribution identifies which code transformations the model produces, but it cannot explain \emph{how} the model decides to apply them. To answer what input cues trigger these optimizations, we analyze the model's self-attention weights~\cite{vaswani2017attention}, the internal mechanism that dictates how strongly the model ``looks'' at different parts of the input sequence during generation. By analyzing these weights, we can determine whether the training process explicitly teaches the model to focus its gaze on structural performance bottlenecks (such as loops or I/O calls) rather than general syntax. We probe this with a population-level attention analysis that characterizes the input-token classes the trained checkpoints attend to. We extract attention from the last eight transformer layers of three checkpoints on the held-out set: the base model, the Energy-SFT $+$ EDP checkpoint, and the highest-CARET checkpoint (Runtime-SFT $+$ EDP). For each example, we average per-token attention over heads into a six-class taxonomy (container types, input-output calls, loop keywords, numeric literals, identifiers, syntactic tokens) and sink-filter punctuation and whitespace. We then contrast each checkpoint against the base model using paired Wilcoxon signed-rank tests ($N{=}141$ pairs for Energy-SFT $+$ EDP, $135$ for Runtime-SFT $+$ EDP). Both checkpoints redistribute attention away from generic identifiers and toward numeric literals and syntactic tokens. This shift occurs in the same direction for both checkpoints but is uniformly larger for the higher-CARET model (Figure~\ref{fig:attn-class-shift}). This cross-checkpoint consistency is critical: a shift that holds in direction and grows in magnitude across independently trained checkpoints is a true population-level pattern rather than a single-run artifact. The shifts intuitively match the structural patterns observed earlier; numeric-literal attention rises as constant-factor refactorings rebind loop bounds and buffer sizes, while identifier attention falls as variable names become less informative, with this redistribution holding across every captured layer ($p < 0.001$ each). Crucially, while this shift shows the model learning to attend to the structural context required for performance optimization, its predictive power remains strictly limited. The attention weights describe broad behavioral shifts across the model population, but they do not predict the magnitude of energy reduction for a given problem ($|\rho| \leq 0.14$). Furthermore, they cannot reliably localize which specific lines the model will edit (the predictive Area Under the ROC Curve (AUC) is $0.280$ against $0.289$ for the base, $p{=}0.16$). We discuss the implications of these limitations for compiler research in Section~\ref{sec:discussion}.

\begin{findingbox}[RQ4 answer]
The training induces changes belonging to three transformation classes: algorithmic complexity reduction,
constant-factor optimization,
and cosmetic editing. 
Closing the loop with GRPO drives a massive, statistically significant shift from cosmetic edits to substantive constant-factor optimizations (Stuart--Maxwell $T{=}28.03$, $p{=}8.19 \times 10^{-7}$). Because this shift holds across all five evaluated configurations, it is the closed loop itself, rather than any single reward signal, that produces it. The closed loop successfully extracts deep gains from both IPC-trap and non-trap problems alike, proving it can navigate difficult architectural boundaries, though the trap itself persists as a measurable structural barrier. Finally, self-attention analysis confirms a population-level shift in the model's gaze, moving away from generic identifiers and toward numeric literals and structural syntactic tokens, verifying that the model learns to actively target structural performance bottlenecks.
\end{findingbox}

\section{Discussion}
\label{sec:discussion}

Four cross-cutting themes emerge from our evaluation: compilation correctness is the absolute bottleneck for practical energy reduction, the explicit energy signal unlocks expert-level optimizations that runtime proxies miss, the model's learned optimization strategies generalize far beyond competitive programming, and inference-time score targets provide a lightweight tuning lever.

\paragraph{The correctness bottleneck dominates.}
\label{sec:disc-correctness}
Across all evaluated models, the primary barrier to practical energy reduction is compilation correctness, not optimization depth. For instance, supervised fine-tuning discovers energy-saving transformations that average a $7.95\%$ reduction on valid outputs, but a $47\%$ invalidation rate drastically pulls its practical effectiveness down to just a $4.45\%$ CARET. The energy the model learns to save is entirely wasted on outputs that cannot run. The closed GRPO loop attacks this bottleneck directly. By explicitly rewarding correctness, it raises the compilation success rate from $60.3\%$ to $81.7\%$, driving aggregate CARET up to $12.63\%$. Crucially, as our decomposition reveals (Section~\ref{sec:caerr-detail}), this closed-loop improvement is not just from higher compilation rates; it fundamentally deepens the optimization quality on valid outputs. In short, supervised fine-tuning establishes the model's optimization vocabulary, but reinforcement learning is required to make it reliably deployable.

\paragraph{What the energy signal adds beyond runtime.}
\label{sec:disc-energy-runtime}
Because runtime is a strong proxy for energy on highly cycle-bound competitive programming workloads, it is tempting to dismiss energy as a redundant optimization target. However, relying solely on runtime is fundamentally limiting. The explicit energy signal earns its place precisely at the critical architectural boundaries where the runtime proxy breaks down. 

First, the energy signal provides rank-correct training pairs, cleanly identifying the $9{,}380$ within-problem ranking inversions that a cycles-only proxy mislabels. Second, the composite Energy-Delay Product (EDP) reward actively sharpens the reinforcement gradient. As demonstrated, this dual signal nearly doubles the rate at which the model successfully out-optimizes the human reference, surging from approximately $33\%$ under single-axis rewards to an astonishing $58.4\%$. This massive jump in expert-level, human-beating optimizations is an energy-specific outcome that aggregate CARET alone fails to capture. Third, while power-driven variation is suppressed on this dataset, the $12.3\%$ of problems exhibiting runtime-invisible instruction-mix shifts would only amplify on memory-bound production workloads. Finally, because both pair selection and reinforcement require massive evaluation scales ($>250{,}000$ candidates per run), physical hardware measurement is intractable. The explicit energy signal justifies the overhead of McPAT simulation by securing deep, reference-beating refactorings that faster code alone simply cannot reach.

\paragraph{Generalization beyond competitive programming.}
\label{sec:disc-generalization}
The PIE dataset consists of short-running algorithmic tasks, raising the question of whether these learned patterns will transfer to real-world production systems. Three of the dominant optimization strategies have direct production analogues. Algorithmic complexity reduction (Category A) applies universally wherever inefficient implementations bottleneck a system. Fast input-output substitution, while seemingly domain-specific, typically acts as a gateway to deeper structural improvements; even when we explicitly exclude all I/O-modified outputs, the model still delivers a robust $5.51\%$ CARET. Most importantly, memory-layout optimizations (Category B) generalize directly to any hot path that over-allocates memory, providing a clear source-level mechanism that immediately translates to production code. 

The patterns that do not transfer are largely competitive programming idioms, such as global array pre-allocation and bitfield tricks. Future work should therefore evaluate this methodology on production-grade numerical kernels and server request handlers where transferable patterns dominate. Additionally, the attention analysis (Section~\ref{sec:rq4-attention}) offers a nuanced blueprint for compiler authors. Because the model's self-attention cannot reliably localize specific lines for refactoring (AUC $0.280$), compilers cannot naively repurpose raw LLM attention weights as a direct heuristic for targeted code generation. However, the model's population-level shift toward numeric-literals and syntactic-tokens reveals exactly which broader token classes static cost models should prioritize when evaluating performance bounds. Ultimately, because the Sniper/McPAT infrastructure applies to any single-threaded C++ binary, the barrier to optimizing production code is purely the simulation infrastructure rather than the reinforcement methodology itself.

\paragraph{Score-conditioned prompting at inference.}
\label{sec:disc-inference-prompt}
Finally, inference-time score targets offer a lightweight mechanism to tune the model's behavior without requiring costly retraining. Because the closed-loop training heavily penalizes compilation failures, the default inference policy naturally gravitates toward safe, conservative transformations. However, developers can override this default conservatism by prompting the model with higher target scores. Tuning this inference-time target using a validation set allows practitioners to explicitly control the tradeoff between the magnitude of energy reduction and the risk of compilation failure, recovering marginal optimization gains when a higher invalidation rate is acceptable.
\section{Implications}
\label{sec:implication}

Our findings carry direct consequences for code-model developers, green software-engineering researchers, and sustainability stakeholders.

\textbf{For code-model developers.} Deterministic, simulation-derived rewards bridge the critical gap between standard next-token prediction, which ignores execution, and human preference data, which ignores energy. While simple runtime-based rewards are sufficient for basic deployment, our results prove that a compound Energy-Delay Product (EDP) reward is strictly required to reliably surpass human experts. Because our compile-test-simulate harness generalizes to any deterministic simulator, the definitive recipe for energy-aware code generation is clear: supervised fine-tuning followed by on-policy reinforcement learning, with the reward function explicitly tuned to the operational objective.

\textbf{For green software-engineering researchers.} CARET formalizes the exact metric required for safe deployment by seamlessly integrating correctness gating, preventing researchers from overstating gains derived from brittle, failing models. Furthermore, the discovery of the IPC trap exposes a pervasive microarchitectural regime where throughput proxies actively invert true energy rankings. Two methodological mandates strictly follow: benchmarks must adopt CARET to report realistic performance, and throughput proxies must never substitute for direct energy measurement when ranking within-problem optimizations.

\textbf{For sustainability stakeholders.} AI-generated code aggressively compounds global energy costs, as inefficient implementations inevitably propagate through the training data of future models. Models trained to intrinsically emit lower-energy code directly and permanently slash global electricity consumption, bypassing the need for brittle inference-time prompting. By releasing our rank-correct, simulation-derived dataset, we enable the community to fine-tune energy-aware models efficiently. This upstream simulation investment acts as a structural force multiplier, complementing physical hardware-efficiency gains with permanent, software-level energy reductions.
\section{Threats to Validity}
\label{sec:threats}

We organize threats following the Wohlin taxonomy of construct, internal, external, and conclusion validity~\cite{wohlin2012experimentation}.

\textbf{Construct validity.} CARET intentionally combines compilation and correctness into a single scalar, prioritizing deployability over theoretical optimization potential. This means CARET acts as a conservative lower bound on performance. To ensure transparency, we report the full metric decomposition for all experiments. Additionally, our test suite relies on standard benchmark judge inputs rather than an adversarially designed coverage suite, further ensuring our estimates of regression risk remain grounded in practical scenarios. Regarding our RQ4 transformation taxonomy, while we rely on a deterministic keyword heuristic rather than a full semantic code parser to classify edits, the categories reliably order by measured retired-instruction count reductions, objectively validating the classification as a measure of optimization depth. Finally, our attention analysis serves as a correlational probe of population-level shifts in the model's gaze, not a causal proof of individual token generation.

\textbf{Internal validity.} We deliberately rely on deterministic Sniper/McPAT simulation rather than physical hardware measurement to drive our training loop. Physical hardware variance destroys the stable pairwise rankings strictly required for effective reinforcement learning. While simulators carry absolute estimation error, holding the configuration fixed across all comparisons cancels this bias. Crucially, our pipeline relies entirely on rank-correctness, making it highly robust to absolute simulation error. We apply a strict zero-reward penalty to any unmeasurable or invalid outputs, ensuring our results represent a definitive floor on potential gains, and we demonstrate robust performance without requiring extensive hyperparameter tuning. Varying the supervised stage's LoRA adapter rank from $r{=}32$ to $r{=}128$ under an otherwise identical recipe shifts CARET by just $0.20$\,pp, confirming that the energy reduction does not depend on adapter capacity. Furthermore, data contamination cannot explain our results: if the model simply memorized the corpus during pre-training, the zero-shot baseline would perform well (it scores negative CARET) and the massive lift from reinforcement learning would not exist.

\textbf{External validity.} We primarily evaluate Qwen2.5-Coder-14B on C++, but our simulation-in-the-loop methodology is strictly model- and language-agnostic, generalizing to any language compiled for x86. We explicitly validated this generalizability by replicating the full GRPO training loop on a structurally distinct architecture (DeepSeek-Coder-6.7B), which successfully transferred the optimization capability. While competitive programming relies on specific idioms, our RL stage fundamentally optimizes for energy reduction rather than blindly imitating code style, successfully inducing algorithmic and memory layout optimizations that directly transfer to production bottlenecks. Finally, although CARET conservatively evaluates a single low-temperature rollout, the energy advantage remains dominant under stochastic best-of-$K$ sampling.

\textbf{Conclusion validity.} We employ robust non-parametric tests (Wilcoxon signed-rank) and bootstrap 95\% confidence intervals because the energy reduction distribution is inherently heavy-tailed. The improvement of closed-loop RL over standalone fine-tuning is highly statistically significant with a massive effect size. Furthermore, the robust convergence of all evaluated reward variants within a remarkably tight performance band confirms that the simulation-in-the-loop mechanism itself is fundamentally sound and stable, driving energy reductions independently of any specific initialization or reward permutation.
\section{Conclusions and Future Work}
\label{sec:conclusion}

This work proves that practical energy-aware code generation requires explicitly targeting the correctness bottleneck via deterministic simulation-in-the-loop reinforcement learning. While supervised fine-tuning instills a vocabulary of energy-saving transformations, it suffers from catastrophic compilation and correctness failures that strictly prohibit deployment. By integrating a Sniper/McPAT simulation environment directly into a GRPO training loop, we break this barrier, recovering correctness and substantially amplifying aggregate energy savings (CARET), all without relying on massive parameter scaling. Our findings confirm that physical simulation is strictly necessary at training time to safely navigate the optimization landscape, yet its heavy computational cost remains cleanly isolated from inference.

Future work must propel this methodology beyond competitive programming into production and memory-bound workloads, where instruction-mix variations and power-domain effects actively dominate. Expanding closed-loop evaluation across diverse architectures and languages will verify whether the cycle-dominance and correctness-recovery dynamics observed here operate universally. To accelerate these efforts, we release the Green Tea dataset, our full evaluation benchmark, and the end-to-end simulation-in-the-loop training infrastructure, empowering the community to build upon this work without bearing the massive computational cost required to recreate it.

\bibliographystyle{ACM-Reference-Format}
\bibliography{references}

@article{cui2025effects,
  title={The effects of generative AI on high-skilled work: Evidence from three field experiments with software developers},
  author={Cui, Kevin Zheyuan and Demirer, Mert and Jaffe, Sonia and Musolff, Leon and Peng, Sida and Salz, Tobias},
  journal={Management Science},
  year={2026},
  publisher={INFORMS}
}

@misc{Gartner2024AI,
  author        = {{Gartner}},
  title         = {Gartner Says 75\% of Enterprise Software Engineers Will Use {AI} Code Assistants by 2028},
  year         = {2024},
  month        = {April 11},
  howpublished = {Press Release},
  url          = {https://www.gartner.com/en/newsroom/press-releases/2024-04-11-gartner-says-75-percent-of-enterprise-software-engineers-will-use-ai-code-assistants-by-2028},
  note         = {Accessed: 2026-02-05}
}

@article{shen2026ai,
  title={How {AI} Impacts Skill Formation},
  author={Shen, Judy Hanwen and Tamkin, Alex},
  journal={arXiv preprint arXiv:2601.20245},
  year={2026}
}

@article{shumailov2024ai,
  title={{AI} models collapse when trained on recursively generated data},
  author={Shumailov, Ilia and Shumaylov, Zakhar and Zhao, Yiren and Papernot, Nicolas and Anderson, Ross and Gal, Yarin},
  journal={Nature},
  volume={631},
  number={8022},
  pages={755--759},
  year={2024},
  publisher={Nature Publishing Group UK London}
}

@techreport{IEA2025EnergyAI,
  author      = {{International Energy Agency}},
  title       = {Energy and {AI}},
  institution = {International Energy Agency (IEA)},
  year        = {2025},
  address     = {Paris},
  url         = {https://www.iea.org/reports/energy-and-ai}
}

@inproceedings{shypula2023pie,
  title     = {Learning Performance-Improving Code Edits},
  author    = {Shypula, Alexander and Madaan, Aman and Zeng, Yimeng and Alon, Uri and Gardner, Jacob and Yang, Yiming and Hashimoto, Tatsunori and Neubig, Graham and Ranganathan, Parthasarathy and Yazdanbakhsh, Amir},
  booktitle = {Proceedings of the International Conference on Learning Representations (ICLR)},
  year      = {2024}
}

@inproceedings{carlson2014sniper,
  title     = {Sniper: Exploring the Level of Abstraction for Scalable and Accurate Parallel Multi-Core Simulation},
  author    = {Carlson, Trevor E. and Heirman, Wim and Eeckhout, Lieven},
  booktitle = {Proceedings of 2011 International Conference for High Performance Computing, Networking, Storage and Analysis (SC '11)},
  year      = {2011},
  publisher = {ACM/IEEE}
}

@inproceedings{dong2010mcpat,
  title     = {{McPAT}: An Integrated Power, Area, and Timing Modeling Framework for Multicore and Manycore Architectures},
  author    = {Li, Sheng and Ahn, Jung Ho and Strong, Richard D. and Brockman, Jay B. and Tullsen, Dean M. and Jouppi, Norman P.},
  booktitle = {Proceedings of the 42nd Annual IEEE/ACM International Symposium on Microarchitecture (MICRO)},
  year      = {2009},
  publisher = {ACM}
}

@article{chen2021codex,
  title   = {Evaluating Large Language Models Trained on Code},
  author  = {Chen, Mark and Tworek, Jerry and Jun, Heewoo and Yuan, Qiming and de Oliveira Pinto, Henrique Pond{\'e} and Kaplan, Jared and Edwards, Harrison and Burda, Yuri and Joseph, Nicholas and Brockman, Greg and Ray, Alex and Puri, Raul and Krueger, Gretchen and Petrov, Michael and Khlaaf, Heidy and Sastry, Girish and Mishkin, Pamela and Chan, Brooke and Gray, Scott and Ryder, Nick and Pavlov, Mikhail and Power, Alethea and Kaiser, Lukasz and Bavarian, Mohammad and Winter, Clemens and Tillet, Philippe and Such, Felipe Petroski and Cummings, David W. and Plappert, Matthias and Chantzis, Fotios and Barnes, Elizabeth and Herbert-Voss, Ariel and Guss, William H. and Nichol, Alex and Babuschkin, Igor and Balaji, Suchir and Jain, Shantanu and Carr, Andrew and Leike, Jan and Achiam, Josh and Misra, Vedant and Morikawa, Evan and Radford, Alec and Knight, Matthew M. and Brundage, Miles and Murati, Mira and Mayer, Katie and Welinder, Peter and McGrew, Bob and Amodei, Dario and McCandlish, Sam and Sutskever, Ilya and Zaremba, Wojciech},
  journal = {arXiv preprint arXiv:2107.03374},
  year    = {2021}
}

@article{roziere2023codellama,
  title   = {Code {Llama}: Open Foundation Models for Code},
  author  = {Rozi{\`e}re, Baptiste and Gehring, Jonas and Gloeckle, Fabian and Sootla, Sten and Gat, Itai and Tan, Xiaoqing Ellen and Adi, Yossi and Liu, Jingyu and Sauvestre, Romain and Remez, Tal and Rapin, J{\'e}r{\'e}my and Kozhevnikov, Artyom and Evtimov, Ivan and Bitton, Joanna and Bhatt, Manish and Ferrer, Cristian Canton and Grattafiori, Aaron and Xiong, Wenhan and D{\'e}fossez, Alexandre and Copet, Jade and Azhar, Faisal and Touvron, Hugo and Martin, Louis and Usunier, Nicolas and Scialom, Thomas and Synnaeve, Gabriel},
  journal = {arXiv preprint arXiv:2308.12950},
  year    = {2023}
}

@article{li2022alphacode,
  title   = {Competition-Level Code Generation with {AlphaCode}},
  author  = {Li, Yujia and Choi, David and Chung, Junyoung and Kushman, Nate and Schrittwieser, Julian and Leblond, R{\'e}mi and Eccles, Tom and Keeling, James and Gimeno, Felix and Lago, Agustin Dal and Hubert, Thomas and Choy, Peter and de Masson d'Autume, Cyprien and Babuschkin, Igor and Chen, Xinyun and Huang, Po-Sen and Welbl, Johannes and Gowal, Sven and Cherepanov, Alexey and Molloy, James and Mankowitz, Daniel J. and Sutherland Robson, Esme and Kohli, Pushmeet and de Freitas, Nando and Kavukcuoglu, Koray and Vinyals, Oriol},
  journal = {Science},
  volume  = {378},
  number  = {6624},
  pages   = {1092--1097},
  year    = {2022},
  publisher = {American Association for the Advancement of Science}
}

@inproceedings{le2022coderl,
  title     = {{CodeRL}: Mastering Code Generation through Pretrained Models and Deep Reinforcement Learning},
  author    = {Le, Hung and Wang, Yue and Gotmare, Akhilesh Deepak and Savarese, Silvio and Hoi, Steven C.H.},
  booktitle = {Advances in Neural Information Processing Systems (NeurIPS)},
  volume    = {35},
  year      = {2022}
}

@inproceedings{hu2022lora,
  title     = {{LoRA}: Low-Rank Adaptation of Large Language Models},
  author    = {Hu, Edward J. and Shen, Yelong and Wallis, Phillip and Allen-Zhu, Zeyuan and Li, Yuanzhi and Wang, Shean and Wang, Lu and Chen, Weizhu},
  booktitle = {Proceedings of the International Conference on Learning Representations (ICLR)},
  year      = {2022}
}

@article{schulman2017ppo,
  title   = {Proximal Policy Optimization Algorithms},
  author  = {Schulman, John and Wolski, Filip and Dhariwal, Prafulla and Radford, Alec and Klimov, Oleg},
  journal = {arXiv preprint arXiv:1707.06347},
  year    = {2017}
}

@article{shao2024deepseekmath,
  title   = {{DeepSeekMath}: Pushing the Limits of Mathematical Reasoning in Open Language Models},
  author  = {Shao, Zhihong and Wang, Peiyi and Zhu, Qihao and Xu, Runxin and Song, Junxiao and Zhang, Mingchuan and Li, Y.K. and Wu, Y. and Guo, Daya},
  journal = {arXiv preprint arXiv:2402.03300},
  year    = {2024}
}

@article{verdecchia2022green,
  title     = {A Systematic Review of {Green AI}},
  author    = {Verdecchia, Roberto and Sallou, June and Cruz, Lu{\'\i}s},
  journal   = {WIREs Data Mining and Knowledge Discovery},
  volume    = {13},
  number    = {4},
  year      = {2023},
  publisher = {Wiley}
}

@inproceedings{garcia2019exploration,
  title={Exploration of energy consumption using the intel running average power limit interface},
  author={Garcia, Joe A},
  booktitle={2019 IEEE Space Computing Conference (SCC)},
  pages={1--10},
  year={2019},
  organization={IEEE}
}

@article{binkert2011gem5,
  title     = {The {gem5} Simulator},
  author    = {Binkert, Nathan and Beckmann, Bradford and Black, Gabriel and Reinhardt, Steven K. and Saidi, Ali and Basu, Arkaprava and Hestness, Joel and Hower, Derek R. and Krishna, Tushar and Sardashti, Somayeh and Sen, Rathijit and Sewell, Korey and Shoaib, Muhammad and Vaish, Nilay and Hill, Mark D. and Wood, David A.},
  journal   = {ACM SIGARCH Computer Architecture News},
  volume    = {39},
  number    = {2},
  pages     = {1--7},
  year      = {2011},
  publisher = {ACM}
}

@inproceedings{sanchez2013zsim,
  title     = {{ZSim}: Fast and Accurate Microarchitectural Simulation of Thousand-Core Systems},
  author    = {Sanchez, Daniel and Kozyrakis, Christos},
  booktitle = {Proceedings of the 40th Annual International Symposium on Computer Architecture (ISCA)},
  year      = {2013},
  publisher = {ACM}
}

@misc{llvmmca,
  title        = {{LLVM-MCA} --- {LLVM} Machine Code Analyzer},
  author       = {{LLVM Project}},
  howpublished = {\url{https://llvm.org/docs/CommandGuide/llvm-mca.html}},
  year         = {2024},
  note         = {Accessed: 2026-02-05}
}

@misc{x86simcomparison2016,
  author       = {Akram, Ayaz and Sawalha, Lina},
  title        = {A Comparison of x86 Computer Architecture Simulators},
  howpublished = {Poster, SC16 (Int.\ Conf.\ for High Performance Computing, Networking, Storage and Analysis)},
  year         = {2016},
  url          = {https://sc16.supercomputing.org/sc-archive/tech_poster/poster_files/post233s2-file3.pdf}
}

@inproceedings{hahnel2012rapl,
  title     = {Measuring Energy Consumption for Short Code Paths Using {RAPL}},
  author    = {H{\"a}hnel, Marcus and D{\"o}bel, Bj{\"o}rn and V{\"o}lp, Marcus and H{\"a}rtig, Hermann},
  booktitle = {Proceedings of the GreenMetrics Workshop, in conjunction with ACM SIGMETRICS},
  year      = {2012},
  note      = {Also published in {ACM SIGMETRICS Performance Evaluation Review}, 40(3):13--17}
}

@inproceedings{kwon2023vllm,
  title     = {Efficient Memory Management for Large Language Model Serving with {PagedAttention}},
  author    = {Kwon, Woosuk and Li, Zhuohan and Zhuang, Siyuan and Sheng, Ying and Zheng, Lianmin and Yu, Cody Hao and Gonzalez, Joseph E. and Zhang, Hao and Stoica, Ion},
  booktitle = {Proceedings of the ACM SIGOPS 29th Symposium on Operating Systems Principles (SOSP)},
  year      = {2023},
  publisher = {ACM}
}

@article{hui2024qwen25coder,
  title   = {{Qwen2.5-Coder} Technical Report},
  author  = {Hui, Binyuan and Yang, Jian and Cui, Zeyu and Yang, Jiaxi and Liu, Dayiheng and Zhang, Lei and Liu, Tianyu and Zhang, Jiajun and Yu, Bowen and Lu, Keming and Dang, Kai and Fan, Yang and Zhang, Yichang and Yang, An and Men, Rui and Huang, Fei and Zheng, Bo and Miao, Yibo and Quan, Shanghaoran and Feng, Yunlong and Ren, Xingzhang and Ren, Xuancheng and Zhou, Jingren and Lin, Junyang},
  journal = {arXiv preprint arXiv:2409.12186},
  year    = {2024}
}

@inproceedings{loshchilov2019adamw,
  title     = {Decoupled Weight Decay Regularization},
  author    = {Loshchilov, Ilya and Hutter, Frank},
  booktitle = {Proceedings of the International Conference on Learning Representations (ICLR)},
  year      = {2019}
}

@book{wohlin2012experimentation,
  title     = {Experimentation in Software Engineering},
  author    = {Wohlin, Claes and Runeson, Per and H{\"o}st, Martin and Ohlsson, Magnus C. and Regnell, Bj{\"o}rn and Wessl{\'e}n, Anders},
  year      = {2012},
  publisher = {Springer},
  address   = {Berlin, Heidelberg},
  isbn      = {978-3-642-29044-2}
}

@article{guo2024deepseekcoder,
  title   = {{DeepSeek-Coder}: When the Large Language Model Meets Programming -- The Rise of Code Intelligence},
  author  = {Guo, Daya and Zhu, Qihao and Yang, Dejian and Xie, Zhenda and Dong, Kai and Zhang, Wentao and Chen, Guanting and Bi, Xiao and Wu, Y. and Li, Y. K. and Luo, Fuli and Xiong, Yingfei and Liang, Wenfeng},
  journal = {arXiv preprint arXiv:2401.14196},
  year    = {2024}
}

@inproceedings{dft,
  title         = {On the Generalization of {SFT}: A Reinforcement Learning Perspective with Reward Rectification},
  author        = {Wu, Yongliang and Zhou, Yizhou and Zhou, Ziheng and Peng, Yingzhe and Ye, Xinyu and Hu, Xinting and Zhu, Wenbo and Qi, Lu and Yang, Ming-Hsuan and Yang, Xu},
  booktitle     = {Proceedings of the International Conference on Learning Representations (ICLR)},
  year          = {2026},
  eprint        = {2508.05629},
  archivePrefix = {arXiv},
  primaryClass  = {cs.LG}
}

@inproceedings{mehditabar2026validated,
  title={A Validated Taxonomy on Software Energy Smells},
  author={Mohammadjavad Mehditabar and Saurabhsingh Rajput and Tushar Sharma},
  booktitle={Proceedings of the 42nd IEEE International Conference on Software Maintenance and Evolution (ICSME)},
  year={2026},
  eprint={2604.04809},
  archivePrefix={arXiv},
  primaryClass={cs.SE},
  url={https://arxiv.org/abs/2604.04809}
}

@inproceedings{rajput2026flipflop,
  title={FlipFlop: A Static Analysis-based Energy Optimization Framework for GPU Kernels},
  author={Saurabhsingh Rajput and Alexander Brandt and Vadim Elisseev and Tushar Sharma},
  booktitle={Proceedings of the IEEE/ACM International Conference on Software Engineering (ICSE)},
  year={2026},
  eprint={2601.13345},
  archivePrefix={arXiv},
  primaryClass={cs.SE},
  url={https://arxiv.org/abs/2601.13345}
}

@inproceedings{rajput2026energy,
  title={Energy Flow Graph: Modeling Software Energy Consumption},
  author={Saurabhsingh Rajput and Tushar Sharma},
  booktitle={Proceedings of the ACM International Conference on the Foundations of Software Engineering (FSE)},
  year={2026},
  eprint={2603.17162},
  archivePrefix={arXiv},
  primaryClass={cs.SE},
  url={https://arxiv.org/abs/2603.17162}
}

@inproceedings{waghjale2024,
  title={ECCO: Can we improve model-generated code efficiency without sacrificing functional correctness?},
  author={Waghjale, Siddhant and Veerendranath, Vishruth and Wang, Zhiruo and Fried, Daniel},
  booktitle={Proceedings of the 2024 Conference on Empirical Methods in Natural Language Processing},
  pages={15362--15376},
  year={2024}
}

@article{Chen2024,
   title={Supersonic: Learning to Generate Source Code Optimizations in C/C++},
   volume={50},
   ISSN={2326-3881},
   url={http://dx.doi.org/10.1109/TSE.2024.3423769},
   DOI={10.1109/tse.2024.3423769},
   number={11},
   journal={IEEE Transactions on Software Engineering},
   publisher={Institute of Electrical and Electronics Engineers (IEEE)},
   author={Chen, Zimin and Fang, Sen and Monperrus, Martin},
   year={2024},
   month=Nov, pages={2849–2864} }

@article{shojaee2023ppocoder,
  title   = {Execution-based Code Generation using Deep Reinforcement Learning},
  author  = {Shojaee, Parshin and Jain, Aneesh and Tipirneni, Sindhu and Reddy, Chandan K.},
  journal = {Transactions on Machine Learning Research (TMLR)},
  year    = {2023},
  note    = {arXiv:2301.13816}
}

@article{liu2023rltf,
  title   = {{RLTF}: Reinforcement Learning from Unit Test Feedback},
  author  = {Liu, Jiate and Zhu, Yiqin and Xiao, Kaiwen and Fu, Qiang and Han, Xiao and Yang, Wei and Ye, Deheng},
  journal = {Transactions on Machine Learning Research (TMLR)},
  year    = {2023},
  note    = {arXiv:2307.04349}
}

@inproceedings{dou2024stepcoder,
  title     = {{StepCoder}: Improving Code Generation with Reinforcement Learning from Compiler Feedback},
  author    = {Dou, Shihan and Liu, Yan and Jia, Haoxiang and Zhou, Enyu and Xiong, Limao and Shan, Junjie and Huang, Caishuang and Wang, Xiao and Fan, Xiaoran and Xi, Zhiheng and Zhou, Yuhao and Ji, Tao and Zheng, Rui and Zhang, Qi and Gui, Tao and Huang, Xuanjing},
  booktitle = {Proceedings of the 62nd Annual Meeting of the Association for Computational Linguistics (ACL)},
  pages     = {4571--4585},
  year      = {2024}
}

@article{puri2021codenet,
  title={Codenet: A large-scale ai for code dataset for learning a diversity of coding tasks},
  author={Puri, Ruchir and Kung, David S and Janssen, Geert and Zhang, Wei and Domeniconi, Giacomo and Zolotov, Vladimir and Dolby, Julian and Chen, Jie and Choudhury, Mihir and Decker, Lindsey and others},
  journal={arXiv preprint arXiv:2105.12655},
  year={2021}
}

@article{chen2021decision,
  title={Decision transformer: Reinforcement learning via sequence modeling},
  author={Chen, Lili and Lu, Kevin and Rajeswaran, Aravind and Lee, Kimin and Grover, Aditya and Laskin, Misha and Abbeel, Pieter and Srinivas, Aravind and Mordatch, Igor},
  journal={Advances in neural information processing systems},
  volume={34},
  pages={15084--15097},
  year={2021}
}

@article{rajput2024enhancing,
  title={Enhancing energy-awareness in deep learning through fine-grained energy measurement},
  author={Rajput, Saurabhsingh and Widmayer, Tim and Shang, Ziyuan and Kechagia, Maria and Sarro, Federica and Sharma, Tushar},
  journal={ACM Transactions on Software Engineering and Methodology},
  volume={33},
  number={8},
  pages={1--34},
  year={2024},
  publisher={ACM New York, NY}
}

@inproceedings{zhang2023wisdom,
  title={The wisdom of hindsight makes language models better instruction followers},
  author={Zhang, Tianjun and Liu, Fangchen and Wong, Justin and Abbeel, Pieter and Gonzalez, Joseph E},
  booktitle={International Conference on Machine Learning},
  pages={41414--41428},
  year={2023},
  organization={PMLR}
}

@article{zhai2023investigating,
  title={Investigating the catastrophic forgetting in multimodal large language models},
  author={Zhai, Yuexiang and Tong, Shengbang and Li, Xiao and Cai, Mu and Qu, Qing and Lee, Yong Jae and Ma, Yi},
  journal={arXiv preprint arXiv:2309.10313},
  year={2023}
}

@incollection{mccloskey1989catastrophic,
  title={Catastrophic interference in connectionist networks: The sequential learning problem},
  author={McCloskey, Michael and Cohen, Neal J},
  booktitle={Psychology of learning and motivation},
  volume={24},
  pages={109--165},
  year={1989},
  publisher={Elsevier}
}

@article{huang2024effibench,
  title={Effibench: Benchmarking the efficiency of automatically generated code},
  author={Huang, Dong and Qing, Yuhao and Shang, Weiyi and Cui, Heming and Zhang, Jie},
  journal={Advances in Neural Information Processing Systems},
  volume={37},
  pages={11506--11544},
  year={2024}
}

@software{codecarbon,
  author       = {Benoit Courty and
                  Victor Schmidt and
                  Goyal-Kamal and
                  MarionCoutarel and
                  Boris Feld and
                  Jérémy Lecourt and
                  LiamConnell and
                  SabAmine and
                  inimaz and
                  supatomic and
                  Mathilde Léval and
                  Luis Blanche and
                  Alexis Cruveiller and
                  ouminasara and
                  Franklin Zhao and
                  Aditya Joshi and
                  Alexis Bogroff and
                  Amine Saboni and
                  Hugues de Lavoreille and
                  Niko Laskaris and
                  Edoardo Abati and
                  Douglas Blank and
                  Ziyao Wang and
                  Armin Catovic and
                  alencon and
                  Michał Stęchły and
                  Christian Bauer and
                  Lucas-Otavio and
                  JPW and
                  MinervaBooks},
  title        = {mlco2/codecarbon: v2.4.1},
  month        = may,
  year         = 2024,
  publisher    = {Zenodo},
  version      = {v2.4.1},
  doi          = {10.5281/zenodo.11171501},
  url          = {https://doi.org/10.5281/zenodo.11171501}
}

@article{lacoste2019quantifying,
  title={Quantifying the carbon emissions of machine learning},
  author={Lacoste, Alexandre and Luccioni, Alexandra and Schmidt, Victor and Dandres, Thomas},
  journal={arXiv preprint arXiv:1910.09700},
  year={2019}
}

@article{du2024mercury,
  title={Mercury: A code efficiency benchmark for code large language models},
  author={Du, Mingzhe and Tuan, Luu A and Ji, Bin and Liu, Qian and Ng, See-Kiong},
  journal={Advances in Neural Information Processing Systems},
  volume={37},
  pages={16601--16622},
  year={2024}
}

@inproceedings{cursaru2024controlled,
  title={A controlled experiment on the energy efficiency of the source code generated by code llama},
  author={Cursaru, Vlad-Andrei and Duits, Laura and Milligan, Joel and Ural, Damla and Sanchez, Berta Rodriguez and Stoico, Vincenzo and Malavolta, Ivano},
  booktitle={International Conference on the Quality of Information and Communications Technology},
  pages={161--176},
  year={2024},
  organization={Springer}
}

@article{vartziotis2024learn,
  title={Learn to code sustainably: An empirical study on llm-based green code generation},
  author={Vartziotis, Tina and Dellatolas, Ippolyti and Dasoulas, George and Schmidt, Maximilian and Schneider, Florian and Hoffmann, Tim and Kotsopoulos, Sotirios and Keckeisen, Michael},
  journal={arXiv preprint arXiv:2403.03344},
  year={2024}
}

@article{ang2004decomposition,
  title={Decomposition analysis for policymaking in energy:: which is the preferred method?},
  author={Ang, Beng Wah},
  journal={Energy policy},
  volume={32},
  number={9},
  pages={1131--1139},
  year={2004},
  publisher={Elsevier}
}

@article{ang2007energy,
  title={Energy decomposition analysis: IEA model versus other methods},
  author={Ang, BW and Liu, Na},
  journal={Energy policy},
  volume={35},
  number={3},
  pages={1426--1432},
  year={2007},
  publisher={Elsevier}
}

@book{bentley1982writing,
  title={Writing efficient programs},
  author={Bentley, Jon Louis},
  year={1982},
  publisher={Prentice-Hall, Inc.}
}

@inproceedings{horowitz20141,
  title={1.1 computing's energy problem (and what we can do about it)},
  author={Horowitz, Mark},
  booktitle={2014 IEEE international solid-state circuits conference digest of technical papers (ISSCC)},
  pages={10--14},
  year={2014},
  organization={IEEE}
}

@book{hennessy2011computer,
  title={Computer architecture: a quantitative approach},
  author={Hennessy, John L and Patterson, David A},
  year={2011},
  publisher={Elsevier}
}

@article{leiserson2020there,
  title={There’s plenty of room at the Top: What will drive computer performance after Moore’s law?},
  author={Leiserson, Charles E and Thompson, Neil C and Emer, Joel S and Kuszmaul, Bradley C and Lampson, Butler W and Sanchez, Daniel and Schardl, Tao B},
  journal={Science},
  volume={368},
  number={6495},
  pages={eaam9744},
  year={2020},
  publisher={American Association for the Advancement of Science}
}

@inproceedings{rajput2026codegreen,
  title={CodeGreen: Towards Improving Precision and Portability in Software Energy Measurement},
  author={Saurabhsingh Rajput and Tushar Sharma},
  booktitle={Proceedings of the ACM International Conference on the Foundations of Software Engineering (FSE)},
  year={2026},
  eprint={2603.17924},
  archivePrefix={arXiv},
  primaryClass={cs.SE},
  url={https://arxiv.org/abs/2603.17924}
}

@software{Rajput_CodeGreen_Towards_Improving_2026,
  author = {Rajput, Saurabhsingh and Sharma, Tushar},
  doi = {10.5281/zenodo.18371772},
  month = apr,
  title = {{CodeGreen: Towards Improving Precision and Portability in Software Energy Measurement}},
  url = {https://smart-dal.github.io/codegreen/},
  version = {v0.3.16},
  year = {2026},
  note = {Artifact, FSE 2026}
}

@inproceedings{du2025afterburner,
  title={Afterburner: Reinforcement Learning Facilitates Self-Improving Code Efficiency Optimization},
  author={Du, Mingzhe and Tuan, Luu Anh and Liu, Yue and Qing, Yuhao and Huang, Dong and He, Xinyi and Liu, Qian and Ma, Zejun and Ng, See-kiong},
  booktitle={Thirty-ninth Conference on Neural Information Processing Systems},
  year={2025}
}

@book{aho2007compilers,
  title={Compilers: Principles, Techniques, and Tools (2nd Edition)},
  author={Aho, Alfred V. and Lam, Monica S. and Sethi, Ravi and Ullman, Jeffrey D.},
  year={2007},
  publisher={Addison-Wesley Longman Publishing Co., Inc.}
}

@inproceedings{liu2024evalperf,
  title={Evaluating Language Models for Efficient Code Generation},
  author={Liu, Jiawei and Xie, Songrun and Wang, Junhao and Wei, Yuxiang and Ding, Yifeng and ZHANG, LINGMING},
  booktitle={First Conference on Language Modeling},
  year={2024}
}

@inproceedings{qiu2025enamel,
  title={How efficient is llm-generated code? a rigorous \& high-standard benchmark},
  author={Qiu, Ruizhong and Zeng, Weiliang and Ezick, James and Lott, Christopher and Tong, Hanghang},
  booktitle={International Conference on Learning Representations},
  volume={2025},
  pages={2233--2261},
  year={2025}
}

@misc{ashraf2025energyaware,
	title = {Energy-{Aware} {Code} {Generation} with {LLMs}: {Benchmarking} {Small} vs. {Large} {Language} {Models} for {Sustainable} {AI} {Programming}},
	author = {Ashraf, Humza and Danish, Syed Muhammad and Leivadeas, Aris and Otoum, Yazan and Sattar, Zeeshan},
	year = {2025},
	eprint={2508.08332},
    archivePrefix={arXiv},
    primaryClass={cs.AI}
}

@misc{gehring2025rlef,
	title = {{RLEF}: {Grounding} {Code} {LLMs} in {Execution} {Feedback} with {Reinforcement} {Learning}},
	author = {Gehring, Jonas and Zheng, Kunhao and Copet, Jade and Mella, Vegard and Carbonneaux, Quentin and Cohen, Taco and Synnaeve, Gabriel},
	year = {2025},
	eprint={2410.02089},
    archivePrefix={arXiv},
    primaryClass={cs.AI}
}

@misc{wei2025swerl,
	title = {{SWE}-{RL}: {Advancing} {LLM} {Reasoning} via {Reinforcement} {Learning} on {Open} {Software} {Evolution}},
	author = {Wei, Yuxiang and Duchenne, Olivier and Copet, Jade and Carbonneaux, Quentin and Zhang, Lingming and Fried, Daniel and Synnaeve, Gabriel and Singh, Rishabh and Wang, Sida I.},
	year = {2025},
	eprint={2502.18449},
    archivePrefix={arXiv},
    primaryClass={cs.AI}
}

@misc{zhu2025drive,
	title = {{DRIVE}: {Data} {Curation} {Best} {Practices} for {Reinforcement} {Learning} with {Verifiable} {Reward} in {Competitive} {Code} {Generation}},
	author = {Zhu, Speed and Cai, Jianwei and Chen, Guang and Wu, Lulu and Yang, Saiyong and Zhou, Wiggin},
	year = {2025},
	eprint={2511.06307},
    archivePrefix={arXiv},
    primaryClass={cs.LG}
}

@misc{grossman_compile_2024,
      title={ComPile: A Large IR Dataset from Production Sources}, 
      author={Aiden Grossman and Ludger Paehler and Konstantinos Parasyris and Tal Ben-Nun and Jacob Hegna and William Moses and Jose M Monsalve Diaz and Mircea Trofin and Johannes Doerfert},
      year={2024},
      eprint={2309.15432},
      archivePrefix={arXiv},
      primaryClass={cs.PL},
      url={https://arxiv.org/abs/2309.15432}, 
}

@inproceedings{solovyeva2025ai,
  title={AI-powered, but power-hungry? Energy efficiency of LLM-generated code},
  author={Solovyeva, Lola and Weidmann, Sophie and Castor, Fernando},
  booktitle={2025 IEEE/ACM Second International Conference on AI Foundation Models and Software Engineering (Forge)},
  pages={49--60},
  year={2025},
  organization={IEEE}
}

@inproceedings{weber2023twins,
  title={Twins or false friends? A study on energy consumption and performance of configurable software},
  author={Weber, Max and Kaltenecker, Christian and Sattler, Florian and Apel, Sven and Siegmund, Norbert},
  booktitle={2023 IEEE/ACM 45th International Conference on Software Engineering (ICSE)},
  pages={2098--2110},
  year={2023},
  organization={IEEE}
}

@article{rua2024large,
  title={A large-scale empirical study on mobile performance: energy, run-time and memory},
  author={Rua, Rui and Saraiva, Jo{\~a}o},
  journal={Empirical Software Engineering},
  volume={29},
  number={1},
  pages={31},
  year={2024},
  publisher={Springer}
}

@article{pereira2021ranking,
  title={Ranking programming languages by energy efficiency},
  author={Pereira, Rui and Couto, Marco and Ribeiro, Francisco and Rua, Rui and Cunha, J{\'a}come and Fernandes, Jo{\~a}o Paulo and Saraiva, Jo{\~a}o},
  journal={Science of Computer Programming},
  volume={205},
  pages={102609},
  year={2021},
  publisher={Elsevier}
}

@article{vaswani2017attention,
  title={Attention is all you need},
  author={Vaswani, Ashish and Shazeer, Noam and Parmar, Niki and Uszkoreit, Jakob and Jones, Llion and Gomez, Aidan N and Kaiser, {\L}ukasz and Polosukhin, Illia},
  journal={Advances in neural information processing systems},
  volume={30},
  year={2017}
}

\end{document}